\documentclass{article}
\usepackage{footnote}
\makesavenoteenv{figure}
\usepackage{arxiv}
\usepackage[font=small]{caption}
\usepackage[utf8]{inputenc} 
\usepackage[T1]{fontenc}    

\usepackage[dvipsnames]{xcolor}
\definecolor{kimiblue}{rgb}{0.09,0.5,0.99}
\usepackage[breaklinks,colorlinks,allcolors=kimiblue]{hyperref}

\usepackage{url}            
\usepackage{booktabs}       
\usepackage{siunitx}        

\usepackage{amsfonts}       
\usepackage{nicefrac}       
\usepackage{microtype}      
\usepackage{lipsum}		
\usepackage{graphicx}
\usepackage{doi}
\usepackage{multirow}
\usepackage{multicol}
\usepackage{xcolor}
\usepackage{threeparttable}
\usepackage{nicematrix}
\usepackage{colortbl}

\usepackage{mathrsfs}
\usepackage[normalem]{ulem}
\usepackage{amsmath,amsfonts,bm}
\usepackage{amssymb,amsthm}
\usepackage{enumitem}
\usepackage{adjustbox}
\usepackage{subcaption}
\usepackage{array}
\usepackage{makecell}

\usepackage{wrapfig}
\usepackage{ulem}
\usepackage[
    backend=biber,
    citestyle=numeric,
    bibstyle=numeric,
    mincitenames=1,
    maxcitenames=1,
    maxbibnames=3,
]{biblatex}

\usepackage{tikz}
\usepackage{color}
\usepackage{threeparttable}
\usepackage{multirow}
\usepackage{multicol}
\usepackage{colortbl}
\usepackage{geometry}
\usepackage{mathtools}
\usepackage{pgfplots}
\usepackage{subcaption}
\usepackage{graphicx}
\usepackage[frozencache,cachedir=_minted]{minted}
\usepackage{placeins}
\usepackage{amsmath}

\usepackage[most]{tcolorbox}
\usepackage[ruled,vlined,linesnumbered]{algorithm2e}
\SetNlSty{texttt}{}{}
\DontPrintSemicolon
\usepackage{fontawesome}

\usetikzlibrary{
  arrows.meta,
  positioning,
  calc,
  shapes.geometric,
  shapes.misc,
  decorations.text,
  patterns,
  patterns.meta,
  fit,
  backgrounds,
  chains,
  shadows,
  math,
  matrix,
  circuits.ee.IEC,
  decorations.pathmorphing,
  decorations.pathreplacing,
  decorations.shapes,
  decorations.pathreplacing,
  calligraphy
}
\usepgfplotslibrary{groupplots}

\definecolor{brickred}{HTML}{b92622}
\definecolor{midnightblue}{HTML}{005c7f}
\definecolor{limegreen}{HTML}{97c65a}
\definecolor{salmon}{HTML}{f1958d}
\definecolor{darkcyan}{HTML}{008B8B}
\definecolor{darkgrey}{rgb}{0.53,0.53,0.53}
\definecolor{mygrey}{rgb}{0.9,0.9,0.9}

\newcommand{\white}[1]{\textcolor{white}{#1}}
\newcommand{\brickred}[1]{\textcolor{brickred}{#1}}

\newcommand{\citep}[1]{\parencite{#1}}
\newcommand{\github}{\raisebox{0pt}{\faGithub}}

\setlist[itemize,1]{leftmargin=\dimexpr 18pt}
\setlist[enumerate,1]{leftmargin=\dimexpr 18pt}

\title{
\raisebox{-0.1\height}{
\includegraphics[width=0.032\textwidth]{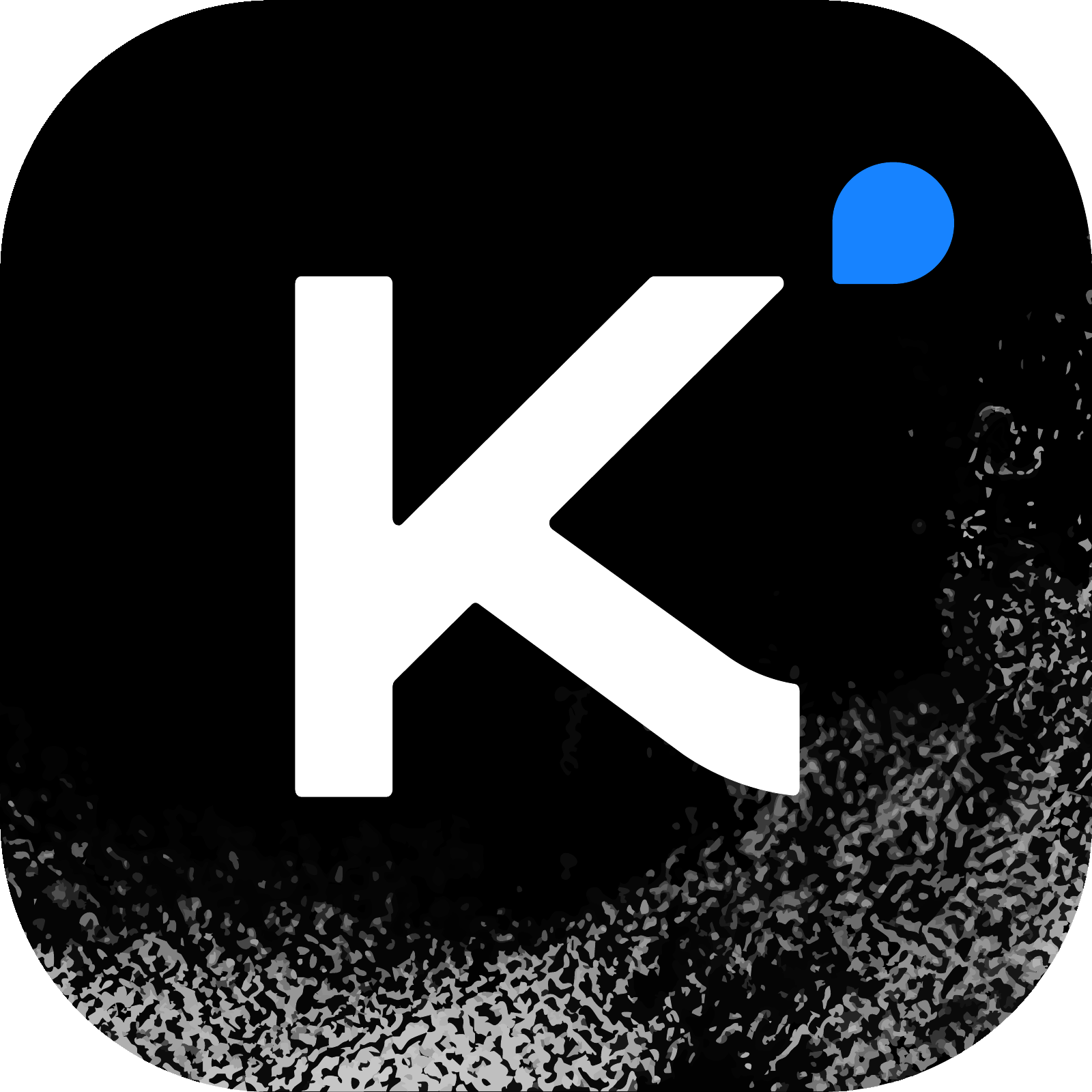}} %
Attention Residuals
}

\author{
Kimi Team\\
\github\,\,\url{https://github.com/MoonshotAI/Attention-Residuals}\\
}

\date{}

\renewcommand{\headeright}{Technical Report}
\renewcommand{\undertitle}{Technical Report of Attention Residuals}
\renewcommand{\shorttitle}{\raisebox{-0.12\height}{\includegraphics[width=0.02\textwidth]{figures/logo.pdf}}
Attention Residuals}

\begin{document}
\maketitle

\vspace{-15pt}
\begin{abstract}
    Residual connections~\citep{he2015resnet} with PreNorm~\citep{xiong2020layer} are standard in modern LLMs, yet they accumulate all layer outputs with fixed unit weights. This uniform aggregation causes uncontrolled hidden-state growth with depth, progressively diluting each layer's contribution~\citep{li2026siamesenorm}.
    We propose \emph{Attention Residuals (AttnRes)}, which replaces this fixed accumulation with $\operatorname{softmax}$ attention over preceding layer outputs, allowing each layer to selectively aggregate earlier representations with learned, input-dependent weights.
    To address the memory and communication overhead of attending over all preceding layer outputs for large-scale model training, we introduce \emph{Block AttnRes}, which partitions layers into blocks and attends over block-level representations, reducing the memory footprint while preserving most of the gains of full AttnRes. Combined with cache-based pipeline communication and a two-phase computation strategy, Block AttnRes becomes a practical drop-in replacement for standard residual connections with minimal overhead.

    Scaling law experiments confirm that the improvement is consistent across model sizes, and ablations validate the benefit of content-dependent depth-wise selection.
    We further integrate AttnRes into the Kimi Linear architecture~\citep{zhang2025kimi} (48B total / 3B activated parameters) and pre-train on 1.4T tokens, where AttnRes mitigates PreNorm dilution, yielding more uniform output magnitudes and gradient distribution across depth, and improves downstream performance across all evaluated tasks.
\end{abstract}
\vspace{-20pt}
\input{figures/model}

\newpage
\section{Introduction}

Standard residual connections~\cite{he2015resnet} are the \emph{de facto} building block of modern LLMs~\cite{openai-2024-gpt4,touvron-2023-llama,deepseekaiv3}. The update $\bm{h}_l = \bm{h}_{l-1} + f_{l-1}(\bm{h}_{l-1})$ is widely understood as a \textit{gradient highway} that lets gradients bypass transformations via identity mappings, enabling stable training at depth. Yet residuals also play a second role that has received less attention. Unrolling the recurrence shows that every layer receives the same uniformly-weighted sum of all prior layer outputs; residuals define how information aggregates across depth. Unlike sequence mixing and expert routing, which now employ learnable input-dependent weighting~\cite{transformer,adaptive,deepseekaiv3}, this depth-wise aggregation remains governed by fixed unit weights, with no mechanism to selectively emphasize or suppress individual layer contributions.

In practice, PreNorm~\cite{xiong2020layer} has become the dominant paradigm, yet its unweighted accumulation causes hidden-state magnitudes to grow as $O(L)$ with depth, progressively diluting each layer's relative contribution~\cite{li2026siamesenorm}. Early-layer information is buried and cannot be selectively retrieved; empirically, a significant fraction of layers can be pruned with minimal loss~\cite{gromov2025unreasonable}. Recent efforts such as scaled residual paths~\cite{wang2022deepnorm} and multi-stream recurrences~\cite{zhu2025hyperconnections} remain bound to the additive recurrence, while methods that do introduce cross-layer access~\cite{pagliardini2024denseformer,xiao2025muddformer} are difficult to scale. The situation parallels the challenges that recurrent neural networks (RNNs) faced over the sequence dimension before attention mechanism provided an alternative.

We observe a formal duality between depth-wise accumulation and the sequential recurrence in RNNs. Building on this duality, we propose \textbf{Attention Residuals (AttnRes)}, which replaces the fixed accumulation $\bm{h}_l = \sum_{i} \bm{v}_i$ with $\bm{h}_l = \sum_{i} \brickred{\alpha_{i \to l}} \cdot \bm{v}_i$, where $\brickred{\alpha_{i \to l}}$ are $\operatorname{softmax}$ attention weights computed from a single learned pseudo-query $\bm{w}_l \in \mathbb{R}^d$ per layer. This lightweight mechanism enables selective, content-aware retrieval across depth with only one $d$-dimensional vector per layer. Indeed, standard residual connections and prior recurrence-based variants can all be shown to perform depth-wise \emph{linear} attention; AttnRes generalizes them to depth-wise $\operatorname{softmax}$ attention, completing for depth the same linear-to-$\operatorname{softmax}$ transition that proved transformative over sequences (\S\ref{sec:structured-depth}, \S\ref{sec:depth-ttt}).

In standard training, Full AttnRes adds negligible overhead, since the layer outputs it requires are already retained for backpropagation. At scale, however, activation recomputation and pipeline parallelism are routinely employed, and these activations must now be explicitly preserved and communicated across pipeline stages. We introduce \textit{Block AttnRes} to maintain efficiency in this regime: layers are partitioned into $N$ blocks, each reduced to a single representation via standard residuals, with cross-block attention applied only over the $N$ block-level summaries. This brings both memory and communication down to $O(Nd)$, and together with infrastructure optimizations (\S\ref{sec:infra}), Block AttnRes serves as a drop-in replacement for standard residual connections with marginal training cost and negligible inference latency overhead.

Scaling law experiments confirm that AttnRes consistently outperforms the baseline across compute budgets, with Block AttnRes matching the loss of a baseline trained with $1.25\times$ more compute. We further integrate AttnRes into the Kimi Linear architecture~\cite{zhang2025kimi} (48B total / 3B activated parameters) and pre-train on 1.4T tokens. Analysis of the resulting training dynamics reveals that AttnRes mitigates PreNorm dilution, with output magnitudes remaining bounded across depth and gradient norms distributing more uniformly across layers. On downstream benchmarks, our final model improves over the baseline across all evaluated tasks.

\textbf{Contributions}

\begin{itemize}[leftmargin=*]
    \item \textbf{Attention Residuals.} We propose AttnRes, which replaces fixed residual accumulation with learned $\operatorname{softmax}$ attention over depth, and its scalable variant Block AttnRes that reduces memory and communication from $O(Ld)$ to $O(Nd)$. Through a unified structured-matrix analysis, we show that standard residuals and prior recurrence-based variants correspond to depth-wise \emph{linear} attention, while AttnRes performs depth-wise $\operatorname{softmax}$ attention.
    \item \textbf{Infrastructure for scale.} We develop system optimizations that make Block AttnRes practical and efficient at scale, including cross-stage caching that eliminates redundant transfers under pipeline parallelism and a two-phase inference strategy that amortizes cross-block attention via online $\operatorname{softmax}$~\citep{milakov2018online}. The resulting training overhead is marginal, and the inference latency overhead is less than 2\% on typical inference workloads.
    \item \textbf{Comprehensive evaluation and analysis.} We validate AttnRes through scaling law experiments, component ablations, and downstream benchmarks on a 48B-parameter model pre-trained on 1.4T tokens, demonstrating consistent improvements over standard residual connections. Training dynamics analysis further reveals that AttnRes mitigates PreNorm dilution, yielding bounded hidden-state magnitudes and more uniform gradient distribution across depth.
\end{itemize}

\newpage

\section{Motivation}

\paragraph{Notation.} Consider a batch of input sequences with shape $B \times T \times d$, where $B$ is the batch size, $T$ is the sequence length, and $d$ is the hidden dimension. For clarity, we write formulas for a single token: $\bm{h}_l \in \mathbb{R}^{d}$ denotes the hidden state entering layer $l$, where $l \in \{1, \ldots, L\}$ is the layer index and $L$ is the total number of layers. The token embedding is $\bm{h}_1$. The function $f_l$ represents the transformation applied by layer $l$. In Transformer models, we treat each self-attention or MLP as an individual \emph{layer}.

\subsection{Training Deep Networks via Residuals}

\paragraph{Residual Learning.} 
Residual learning~\citep{he2015resnet} proves to be a critical technique in training deep networks as it allows gradients to bypass transformations.
Specifically, each layer updates the hidden state as:
$$\bm{h}_{l} = \bm{h}_{l-1} + f_{l-1}(\bm{h}_{l-1})$$
Expanding this recurrence, the hidden state at layer $l$ is the sum of the embedding and all preceding layer outputs: $\bm{h}_l = \bm{h}_1 + \sum_{i=1}^{l-1} f_i(\bm{h}_i)$.
The key insight behind residual connections is \emph{identity mapping}: each layer preserves a direct path for both information and gradients to flow unchanged. During back-propagation, the gradient with respect to an intermediate hidden state is:
$$\frac{\partial \mathcal{L}}{\partial \bm{h}_l} = \frac{\partial \mathcal{L}}{\partial \bm{h}_{L}} \cdot \prod_{j=l}^{L-1} \left( \mathbf{I} + \frac{\partial f_j}{\partial \bm{h}_j} \right)$$
Expanding this product yields $\mathbf{I}$ plus higher-order terms involving the layer Jacobians $\partial f_j/\partial \bm{h}_j$. The identity term is always preserved, providing a direct gradient path from the loss to any layer regardless of depth.

\paragraph{Generalizing Residuals.}
While effective, the fixed unit coefficients in the residual update treat every layer's contribution uniformly, offering no mechanism to adapt the mixing across depth.
Highway networks~\citep{srivastava2015highway} relax this by introducing learned element-wise gates:
$$\bm{h}_{l} = (1 - \bm{g}_{l}) \odot \bm{h}_{l-1} + \bm{g}_{l} \odot f_{l-1}(\bm{h}_{l-1})$$
where $\bm{g}_l \in [0,1]^d$ interpolates between the transformation and the identity path.
More generally, both are instances of a weighted recurrence $\bm{h}_{l} = \alpha_{l} \cdot \bm{h}_{l-1} + \beta_{l} \cdot f_{l-1}(\bm{h}_{l-1})$, with residual setting $\alpha_{l} {=} \beta_{l} {=} 1$ and Highway setting $\alpha_{l} {=} 1 {-} \bm{g}_{l},\, \beta_{l} {=} \bm{g}_{l}$.

\paragraph{Limitations.}
Whether fixed or gated, both approaches share a fundamental constraint: each layer can only access its immediate input $\bm{h}_{l-1}$, a single compressed state that conflates all earlier layer outputs, rather than the individual outputs themselves.
This entails several limitations:
(1)~\emph{no selective access}: different layer types (e.g., attention vs.\ MLP) receive the same aggregated state, despite potentially benefiting from different weightings;
(2)~\emph{irreversible loss}: information lost through aggregation cannot be selectively recovered in deeper layers;
and (3)~\emph{output growth}: later layers learn increasingly larger outputs to gain influence over the accumulated residual, which can destabilize training.
These limitations motivate a mechanism that lets each layer selectively aggregate information from all preceding layers.

\section{Attention Residuals: A Unified View of Time and Depth}
\label{sec:attnres}

The limitations discussed above are reminiscent of similar bottlenecks in sequence modeling, suggesting that we seek similar solutions for the depth dimension.

\paragraph{The Duality of Time and Depth.}
Like RNNs over time, residual connections compress all prior information into a single state $\bm{h}_l$ over depth.
For sequence modeling, the Transformer improved upon RNNs by replacing recurrence with attention~\citep{bahdanau2016attention,vaswani-2017-attention}, allowing each position to selectively access all previous positions with data-dependent weights.
We propose the same methodology for depth:
\begin{equation}
    \bm{h}_{l} = \brickred{\alpha_{0 \to l}} \cdot \bm{h}_1 + \sum_{i=1}^{l-1} \brickred{\alpha_{i \to l}} \cdot f_i(\bm{h}_{i})
\end{equation}
where $\brickred{\alpha_{i \to l}}$ are layer-specific attention weights satisfying $\sum_{i=0}^{l-1} \brickred{\alpha_{i \to l}} = 1$.
Unlike sequence length (which can reach millions of tokens), network depth is typically modest ($L < 1000$), making $O(L^2)$ attention over depth computationally feasible.
We call this approach \emph{Attention Residuals}, abbreviated as \emph{AttnRes}.

\subsection{Full Attention Residuals}
The attention weights can be written as $\brickred{\alpha_{i \to l}} = \phi(\bm{q}_{l},\, \bm{k}_{i})$ for a kernel function $\phi\colon \mathbb{R}^d \times \mathbb{R}^d \to \mathbb{R}_{\geq 0}$, where $\bm{q}_l$ and $\bm{k}_i$ are query and key vectors~\cite{katharopoulos-2020-transformers, zhong2025understanding}. Different choices of $\phi$ recover different residual variants (\S\ref{sec:structured-depth}); we adopt $\phi(\bm{q}, \bm{k}) = \exp\left(\bm{q}^\top\operatorname{RMSNorm}(\bm{k})\right)$~\cite{zhang2019root} with normalization, yielding $\operatorname{softmax}$ attention over depth:
\begin{equation}
    \label{eq:attn-weights}
    \brickred{\alpha_{i \to l}} = \frac{\phi\left(\bm{q}_{l}, \bm{k}_{i}\right)}{\sum_{j=0}^{l-1} \phi\left(\bm{q}_{l}, \bm{k}_{j}\right)}
\end{equation}

For each layer $l$, we define:
\begin{equation}
    \label{eq:qkvattn}
    \bm{q}_{l} = \bm{w}_{l}, \quad
    \quad \bm{k}_{i} = \bm{v}_{i} = \begin{cases} \bm{h}_1 & i = 0 \\ f_i(\bm{h}_{i}) & 1 \leq i \leq l-1 \end{cases}
\end{equation}
where the query $\bm{q}_l=\bm{w}_{l}$ is a layer-specific learnable vector in $\mathbb{R}^{d}$. The $\operatorname{RMSNorm}$ inside $\phi$ prevents layers with large-magnitude outputs from dominating the attention weights. The input to layer $l$ is then:
\begin{equation}
    \label{eq:full-attn}
    \bm{h}_{l} = \sum_{i=0}^{l-1} \brickred{\alpha_{i \to l}} \cdot \bm{v}_{i}
\end{equation}

We call this form \emph{full attention residuals}. For each token, Full AttnRes requires $O(L^2 d)$ arithmetic and $O(Ld)$ memory to store layer outputs. Since depth is far smaller than sequence length, the arithmetic cost is modest.

\paragraph{Overhead.}
The $O(Ld)$ memory overlaps entirely with the activations already retained for backpropagation, so Full AttnRes introduces no additional memory overhead in vanilla training. At scale, however, activation recomputation and pipeline parallelism are widely adopted: layer outputs that would otherwise be freed and recomputed must now be kept alive for all subsequent layers, and under pipeline parallelism each must further be transmitted across stage boundaries. Both the memory and communication overhead then grow as $O(Ld)$.

\paragraph{Blockwise optimization.}
A deliberate design choice in Full AttnRes is that the \emph{pseudo}-query $\bm{w}_l$ is a learned parameter decoupled from the layer's forward computation. This independence means that attention weights for any group of layers can be computed in parallel without waiting for their sequential outputs, and in particular permits grouping the $L$ layers into $N$ blocks of $S$ layers each and batching the attention computation within each block, reducing per-layer memory I/O from $O(Ld)$ to $O((S{+}N)d)$ (we defer the detailed two-phase strategy to \S\ref{sec:infra}). Under current distributed training regimes, however, the dominant cost is not local memory bandwidth but cross-stage communication under pipeline parallelism: every layer output must still be transmitted between stages, and this $O(Ld)$ communication overhead cannot be alleviated by local batching. This motivates the Block AttnRes variant introduced below, which reduces the number of cross-stage representations from $L$ to $N$. We anticipate that future interconnect improvements will make the full $O(Ld)$ communication practical, fully realizing the potential of Full AttnRes.

\subsection{Block Attention Residuals}

We propose \emph{Block Attention Residuals}, which partitions the $L$ layers into $N$ blocks: within each block, the layer outputs are reduced to a single representation via summation, and across blocks, we apply full attention over only $N$ block-level representations and the token embedding. This reduces both memory and communication overhead from $O(Ld)$ to $O(Nd)$.

\paragraph{Intra-Block Accumulation.}
Specifically, we divide the $L$ layers into $N$ blocks of $S = L/N$ layers each, assuming $L$ is divisible by $N$; otherwise, the last block contains the remaining $L \bmod N$ layers. Let $\mathcal{B}_n$ denote the set of layer indices in block $n$ ($n = 1, \ldots, N$).
To form a block, we sum all of its layer outputs:
\begin{equation}
    \bm{b}_n = \sum_{j \in \mathcal{B}_n} f_j(\bm{h}_j)
\end{equation}
We further denote $\bm{b}_n^i$ as the partial sum over the first $i$ layers in $\mathcal{B}_n$, so that $\bm{b}_n = \bm{b}_n^{S}$.
When $L$ is not divisible by $N$, the final partial sum is taken as the last block's representation.
As in Full AttnRes, the $\operatorname{RMSNorm}$ inside $\phi$  prevents magnitude differences between complete blocks and partial sums from biasing the attention weights.

\paragraph{Inter-Block Attention.}
In Full AttnRes, the input to layer $l$ is computed by attending over all outputs up to $f_{l-1}(\bm{h}_{l-1})$.
The block-wise variant replaces these individual outputs with block representations, defining $\bm{b}_0 = \bm{h}_1$ so that the token embedding is always included as a source.
For the $i$-th layer in block $n$, the value matrix is:
\begin{equation}
    \mathbf{V} = \begin{cases}
        [\bm{b}_0, \bm{b}_1, \ldots, \bm{b}_{n-1}]^\top                 & \text{if } i = 1 \text{ (first layer of block } n\text{)} \\
        [\bm{b}_0, \bm{b}_1, \ldots, \bm{b}_{n-1}, \bm{b}_n^{i-1}]^\top & \text{if } i \geq 2 \text{ (subsequent layers)}           \\
    \end{cases}
\end{equation}
Keys and attention weights follow Eq.~\ref{eq:qkvattn} and Eq.~\ref{eq:attn-weights}.
The input of the very first layer of the network is the token embeddings, i.e. $\bm{b}_0 = \bm{h}_1$. In each block, the first layer receives the previous block representations and the token embeddings, and the subsequent layers additionally attend to the partial sum $\bm{b}_n^{i-1}$. The final output layer aggregates all $N$ block representations.
Fig.~\ref{listing:attnres} provides PyTorch-style pseudocode for Block AttnRes.
\renewcommand{\theFancyVerbLine}{\ttfamily\textcolor[rgb]{0.5,0.5,0.5}{\scriptsize\arabic{FancyVerbLine}}}

\begin{figure}[t!]
    \centering
    \let\textit\textup
    \begin{minted}[
    fontsize=\scriptsize,
    linenos,
    numbersep=6pt,
    xleftmargin=16pt,
    ]{python}
def block_attn_res(blocks: list[Tensor], partial_block: Tensor, proj: Linear, norm: RMSNorm) -> Tensor:
    """
    Inter-block attention: attend over block reps + partial sum.
    blocks:
        N tensors of shape [B, T, D]: completed block representations for each previous block
    partial_block:
        [B, T, D]:    intra-block partial sum (b_n^i)
    """
    V = torch.stack(blocks + [partial_block])  # [N+1, B, T, D]
    K = norm(V)
    logits = torch.einsum('d, n b t d -> n b t', proj.weight.squeeze(), K)
    h = torch.einsum('n b t, n b t d -> b t d', logits.softmax(0), V)
    return h

def forward(self, blocks: list[Tensor], hidden_states: Tensor) -> tuple[list[Tensor], Tensor]:
    partial_block = hidden_states
    # apply block attnres before attn
    # blocks already include token embedding
    h = block_attn_res(blocks, partial_block, self.attn_res_proj, self.attn_res_norm)

    # if reaches block boundary, start new block
    # block_size counts ATTN + MLP; each transformer layer has 2
    if self.layer_number % (self.block_size // 2) == 0:
        blocks.append(partial_block)
        partial_block = None

    # self-attention layer
    attn_out = self.attn(self.attn_norm(h))
    partial_block = partial_block + attn_out if partial_block is not None else attn_out

    # apply block attnres before MLP
    h = block_attn_res(blocks, partial_block, self.mlp_res_proj, self.mlp_res_norm)

    # MLP layer
    mlp_out = self.mlp(self.mlp_norm(h))
    partial_block = partial_block + mlp_out

    return blocks, partial_block
\end{minted}
    \caption{PyTorch-style pseudo code for Block Attention Residuals. \texttt{block\_attn\_res} computes $\operatorname{softmax}$ attention over block representations using a learned pseudo-query $\bm{w}_l$; \texttt{forward} is a single-layer pass that maintains \texttt{partial\_block} ($\bm{b}_n^i$, intra-block residual) and \texttt{blocks} ($[\bm{b}_0, \ldots, \bm{b}_{n-1}]$, inter-block history).}
    \label{listing:attnres}
\end{figure}

\paragraph{Efficiency.}
Since each layer now attends over $N$ block representations rather than $L$ individual outputs, memory reduces from $O(L)$ to $O(N)$ and computation from $O(L^2)$ to $O(N^2)$.
The block count $N$ interpolates between two extremes: $N = L$ recovers Full AttnRes, while $N = 1$ reduces to standard residual connections with the embedding isolated as $\bm{b}_0$.
Empirically, we find that $N \approx 8$ recovers most of the benefit across model scales, requiring only eight stored hidden states per token (see \S~\ref{sec:exp}).

Beyond memory and computation, the block structure also benefits inference latency: block boundaries define the dispatch granularity for the blockwise optimization described in \S\ref{sec:attnres}, and the fixed block count $N$ bounds the KV cache size. The parallel inter-block results are merged with the sequential intra-block partial sums via online $\operatorname{softmax}$~\citep{milakov2018online}, preserving exact equivalence (\S\ref{sec:infra}).
\section{Infrastructure Design}
\label{sec:infra}

Block AttnRes introduces additional system challenges compared to standard residual connections. For large-scale model training, block representations must be propagated across pipeline stages, causing heavy communication in a na\"ive implementation. During inference, repeated access to accumulated block representations increases latency, while long-context prefilling amplifies the memory cost of caching block representations. We address these challenges with cross-stage caching in training, and with a two-phase computation strategy together with a memory-efficient prefilling scheme in inference.

\subsection{Training}
\label{sec:infra-training}
For small-scale training, AttnRes adds a tiny computation overhead and no extra memory usage, as the activations need to be saved for backpropagation regardless.
Under large-scale distributed training, pipeline parallelism poses the primary infrastructure challenge for AttnRes.
Full AttnRes requires all $L$ layer outputs to be transmitted across stages; Block AttnRes reduces this to $N$ block representations, and the optimizations below further minimize the remaining overhead.

\paragraph{Pipeline communication.}
With standard residual connections, pipeline parallelism~\citep{huang2019gpipe} transfers a fixed-size hidden state between adjacent stages, independent of pipeline depth.
Block AttnRes requires all accumulated block representations at each stage for inter-block attention, and na\"ively transmitting the full history at every transition incurs redundant communication.

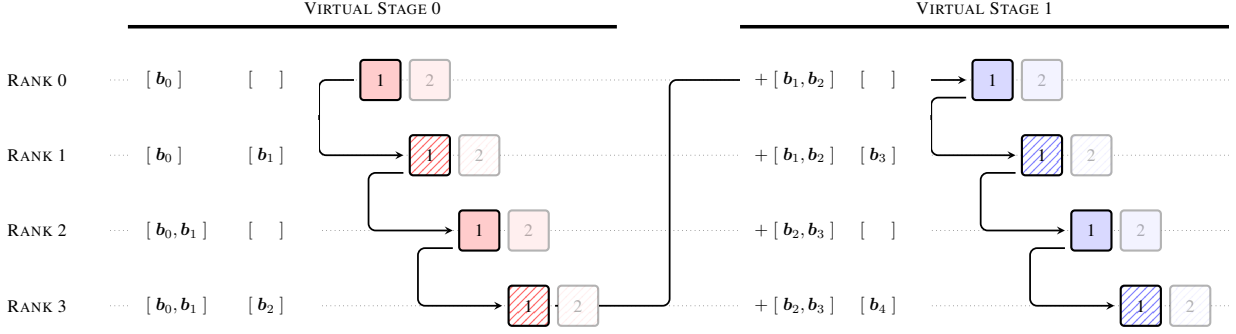
\begin{figure*}[t]
    \centering
    \pgfdeclarelayer{arrowlayer}
    \pgfdeclarelayer{boxlayer}
    \pgfsetlayers{background,arrowlayer,main,boxlayer}
    \begin{adjustbox}{width=0.99\textwidth,center}
        \begin{tikzpicture}[
                box/.style={
                        rectangle,
                        draw,
                        very thick,
                        rounded corners=2pt,
                        minimum width=0.8cm,
                        minimum height=0.8cm,
                        font=\normalsize
                    },
                round0/.style={box, fill=red!20, draw=black},
                round1/.style={box, fill=blue!15, draw=black},
                round3/.style={box, fill=teal!35, draw=black},
                round0pattern/.style={box, preaction={fill=white}, fill=red!20, draw=black, pattern=north east lines, pattern color=red!60},
                round1pattern/.style={box, preaction={fill=white}, fill=blue!15, draw=black, pattern=north east lines, pattern color=blue!50},
                round3pattern/.style={box, preaction={fill=white}, fill=teal!35, draw=black, pattern=north east lines, pattern color=teal!70},
                label/.style={above},
                cache/.style={below},
                rank/.style={anchor=east},
                arrow/.style={->, >=stealth, line width=1pt, rounded corners=4pt, shorten >=3pt, shorten <=3pt}
            ]

            \node[rank] at (-0.5, 0) {\textsc{Rank 0}};
            \node[rank] at (-0.5, -1.5) {\textsc{Rank 1}};
            \node[rank] at (-0.5, -3) {\textsc{Rank 2}};
            \node[rank] at (-0.5, -4.5) {\textsc{Rank 3}};

            \node[fill=white, text width=60pt, inner sep=10pt, align=left] (cache0r1) at (2.0, 0) {$[\;\bm{b}_0\;]$};
            \node[fill=white, text width=25pt, inner sep=15pt, align=left] (cache0br1) at ([xshift=0pt]cache0r1.east) {$[\;\phantom{\bm{b}_0}\;]$};
            \node[fill=white, text width=60pt, inner sep=10pt, align=left] (cache0r2) at (2.0, -1.5) {$[\;\bm{b}_0\;]$};
            \node[fill=white, text width=25pt, inner sep=15pt, align=left] (cache0br2) at ([xshift=0pt]cache0r2.east) {$[\;\bm{b}_1\;]$};
            \node[fill=white, text width=60pt, inner sep=10pt, align=left] (cache0r3) at (2.0, -3) {$[\;\bm{b}_0,\bm{b}_1\;]$};
            \node[fill=white, text width=25pt, inner sep=15pt, align=left] (cache0br3) at ([xshift=0pt]cache0r3.east) {$[\;\phantom{\bm{b}_0}\;]$};
            \node[fill=white, text width=60pt, inner sep=10pt, align=left] (cache0r4) at (2.0, -4.5) {$[\;\bm{b}_0,\bm{b}_1\;]$};
            \node[fill=white, text width=25pt, inner sep=15pt, align=left] (cache0br4) at ([xshift=0pt]cache0r4.east) {$[\;\bm{b}_2\;]$};

            \begin{pgfonlayer}{boxlayer}
                \node[round0] (r0c0) at ([xshift=35pt]cache0br1.east) {1};
                \fill[white] ([xshift=63pt]cache0br1.east) ++(-0.4cm,-0.4cm) rectangle ++(0.8cm,0.8cm);
                \node[round0, opacity=0.3] (r0c0b) at ([xshift=63pt]cache0br1.east) {2};
                \node[round0pattern] (r1c0) at ([xshift=63pt]cache0br2.east) {1};
                \fill[white] ([xshift=91pt]cache0br2.east) ++(-0.4cm,-0.4cm) rectangle ++(0.8cm,0.8cm);
                \node[round0pattern, opacity=0.3] (r1c0b) at ([xshift=91pt]cache0br2.east) {2};
                \node[round0] (r2c0) at ([xshift=91pt]cache0br3.east) {1};
                \fill[white] ([xshift=119pt]cache0br3.east) ++(-0.4cm,-0.4cm) rectangle ++(0.8cm,0.8cm);
                \node[round0, opacity=0.3] (r2c0b) at ([xshift=119pt]cache0br3.east) {2};
                \node[round0pattern] (r3c0) at ([xshift=119pt]cache0br4.east) {1};
                \fill[white] ([xshift=147pt]cache0br4.east) ++(-0.4cm,-0.4cm) rectangle ++(0.8cm,0.8cm);
                \node[round0pattern, opacity=0.3] (r3c0b) at ([xshift=147pt]cache0br4.east) {2};
            \end{pgfonlayer}

            \node[fill=white, text width=60pt, inner sep=10pt, align=left] (cache1r1) at ([xshift=120pt]r3c0b.east |- cache0r1) {$+\; [\;\bm{b}_1, \bm{b}_2\;]$};
            \node[fill=white, text width=25pt, inner sep=15pt, align=left] (cache1br1) at ([xshift=0pt]cache1r1.east) {$[\;\phantom{\bm{b}_3}\;]$};
            \node[fill=white, text width=60pt, inner sep=10pt, align=left] (cache1r2) at ([xshift=120pt]r3c0b.east |- cache0r2) {$+\; [\;\bm{b}_1, \bm{b}_2\;]$};
            \node[fill=white, text width=25pt, inner sep=15pt, align=left] (cache1br2) at ([xshift=0pt]cache1r2.east) {$[\;\bm{b}_3\;]$};
            \node[fill=white, text width=60pt, inner sep=10pt, align=left] (cache1r3) at ([xshift=120pt]r3c0b.east |- cache0r3) {$+\; [\;\bm{b}_2,\bm{b}_3\;]$};
            \node[fill=white, text width=25pt, inner sep=15pt, align=left] (cache1br3) at ([xshift=0pt]cache1r3.east) {$[\;\phantom{\bm{b}_4}\;]$};
            \node[fill=white, text width=60pt, inner sep=10pt, align=left] (cache1r4) at ([xshift=120pt]r3c0b.east |- cache0r4) {$+\; [\;\bm{b}_2,\bm{b}_3\;]$};
            \node[fill=white, text width=25pt, inner sep=15pt, align=left] (cache1br4) at ([xshift=0pt]cache1r4.east) {$[\;\bm{b}_4\;]$};

            \begin{pgfonlayer}{boxlayer}
                \node[round1] (r0c1) at ([xshift=35pt]cache1br1.east) {1};
                \fill[white] ([xshift=63pt]cache1br1.east) ++(-0.4cm,-0.4cm) rectangle ++(0.8cm,0.8cm);
                \node[round1, opacity=0.3] (r0c1b) at ([xshift=63pt]cache1br1.east) {2};
                \node[round1pattern] (r1c1) at ([xshift=63pt]cache1br2.east) {1};
                \fill[white] ([xshift=91pt]cache1br2.east) ++(-0.4cm,-0.4cm) rectangle ++(0.8cm,0.8cm);
                \node[round1pattern, opacity=0.3] (r1c1b) at ([xshift=91pt]cache1br2.east) {2};
                \node[round1] (r2c1) at ([xshift=91pt]cache1br3.east) {1};
                \fill[white] ([xshift=119pt]cache1br3.east) ++(-0.4cm,-0.4cm) rectangle ++(0.8cm,0.8cm);
                \node[round1, opacity=0.3] (r2c1b) at ([xshift=119pt]cache1br3.east) {2};
                \node[round1pattern] (r3c1) at ([xshift=119pt]cache1br4.east) {1};
                \fill[white] ([xshift=147pt]cache1br4.east) ++(-0.4cm,-0.4cm) rectangle ++(0.8cm,0.8cm);
                \node[round1pattern, opacity=0.3] (r3c1b) at ([xshift=147pt]cache1br4.east) {2};
            \end{pgfonlayer}

            \begin{scope}[on background layer]
                \foreach \y in {0, -1.5, -3, -4.5} {
                        \draw[dotted, gray] ([xshift=-10pt]cache0r1.west |- 0,\y) -- ([xshift=10pt]r3c1b.east |- 0,\y);
                    }
            \end{scope}

            \coordinate (round1y) at ([yshift=15pt]cache0r1.north);
            \draw[line width=2pt] (cache0r1.north west |- round1y) -- ([xshift=10pt]r3c0b.north east |- round1y) node[midway, above=3pt] {\textsc{Virtual Stage 0}};
            \coordinate (round2y) at ([yshift=15pt]cache1r1.north);
            \draw[line width=2pt] (cache1r1.north west |- round2y) -- ([xshift=10pt]r3c1b.north east |- round2y) node[midway, above=3pt] {\textsc{Virtual Stage 1}};

            \begin{pgfonlayer}{arrowlayer}
                \draw[arrow] (r0c0.west) -- ++(-0.8,0) |- ([yshift=0pt]r1c0.west);
                \draw[arrow] ([yshift=-10pt]r1c0.west) -- ++(-0.8,0) |- ([yshift=0pt]r2c0.west);
                \draw[arrow] ([yshift=-10pt]r2c0.west) -- ++(-0.8,0) |- (r3c0.west);
                \draw[arrow] (r3c0.east) -- ($(r3c0b.east)!0.5!(cache1r1.west |- r3c0.east)$) |- (r0c1.west);
                \draw[arrow] ([yshift=-10pt]r0c1.west) -- ++(-0.8,0) |-     ([yshift=0pt]r1c1.west);
                \draw[arrow] ([yshift=-10pt]r1c1.west) -- ++(-0.8,0) |- ([yshift=0pt]r2c1.west);
                \draw[arrow] ([yshift=-10pt]r2c1.west) -- ++(-0.8,0) |- (r3c1.west);
            \end{pgfonlayer}

        \end{tikzpicture}
    \end{adjustbox}
    \caption{
    Cache-based pipeline communication example with 4 physical ranks and 2 virtual stages per rank, where hatched boxes denote end of AttnRes blocks.
    Numbers indicate micro-batch indices. Each rank caches previously received blocks; stage transitions only transmit incremental blocks ($+[\bm{b}_1, \bm{b}_2]$) instead of the full history.
    }
    \label{fig:pipeline}
\end{figure*}

Consider an interleaved pipeline schedule~\citep{narayanan2021megatron} with $P$ physical stages and $V$ virtual stages per physical stage.
For simplicity, assume each physical stage produces on average $N_p$ block representations of dimension $d$ per token.\footnote{In practice, block boundaries need not align with physical stage boundaries. For example, in Fig.~\ref{fig:pipeline}, each block spans two physical stages, so only every other transition involves a newly completed block.}
With $C = PV$ total chunks (each physical stage in each virtual stage), the $j$-th chunk accumulates $jN_p$ blocks.
Na\"ively transmitting all accumulated blocks at every transition incurs per-token communication cost:
\begin{equation}
    \mathrm{Comm}_{\text{na\"ive}} = \sum_{j=1}^{C-1} jN_p \cdot d = \frac{C(C{-}1)}{2}\,N_p d.
\end{equation}
\paragraph{Cross-stage caching.}
Since each physical stage processes multiple virtual stages in succession, we can eliminate this redundancy by caching blocks locally: blocks received during earlier virtual stages remain in local memory and need not be re-transmitted.
The first virtual stage ($v = 1$) has no cache and accumulates normally; for $v \geq 2$, each transition conveys only the ${\sim}P N_p$ incremental blocks accumulated since the receiver's corresponding chunk in the previous virtual stage.
Total communication reduces to:
\begin{equation}
    \mathrm{Comm}_{\text{cached}} = \underbrace{\frac{P(P{-}1)}{2}\, N_p d}_{\text{first virtual stage}} + \underbrace{(V{-}1)\, P^2\, N_p d}_{\text{subsequent virtual stages}}.
\end{equation}

Caching reduces peak per-transition cost from $O(C)$ to $O(P)$, a $V{\times}$ improvement that enables full overlap with computation during steady-state 1F1B.
The backward pass benefits from the same scheme. Fig.~\ref{fig:pipeline} illustrates this optimization with $P{=}4$ and $V{=}2$: for the second virtual stage, caching eliminates 6 redundant block transmissions.

\paragraph{Memory overhead.}
With cross-stage caching, each block is stored exactly once across all $V$ virtual stages, which becomes negligible relative to standard per-layer activation cache. Crucially, the per-layer activation footprint remains identical to standard architectures, as activation checkpointing eliminates all inter-block attention intermediates, and the checkpointed input $\bm{p}_l$ matches the memory size of the hidden state $\bm{h}_l$ it replaces.

In terms of wall-clock time, Block AttnRes adds negligible training overhead when pipeline parallelism is not enabled; under pipeline parallelism, the measured end-to-end overhead is less than 4\%.

\subsection{Inference}
\label{sec:infra-inference}
The two-phase computation strategy described below applies to both Full and Block AttnRes: in either case, layers are grouped into blocks of size $S$, with Phase~1 batching the inter-block queries and Phase~2 handling sequential intra-block lookback. For Full AttnRes, this reduces per-layer I/O from $O(Ld)$ to $O((S{+}N)d)$ (detailed derivation shown in Appendix \ref{app:inference}); Block AttnRes further reduces the stored representations from $L$ to $N$, since each block is compressed into a single vector.
In what follows, we focus on Block AttnRes and detail the two-phase computation strategy together with a sequence-sharded prefilling scheme for long-context inputs.

\paragraph{Two-phase computation strategy.}
The layer-wise attention computation of Block AttnRes resembles autoregressive decoding, where block representations serve as a shared KV cache reused across layers.
A na\"ive implementation computes the attention residual at every layer, each requiring a full pass over all preceding blocks, resulting in $O(L \cdot N)$ memory accesses.
Since the pseudo-query vectors are decoupled from the forward computation (\S\ref{sec:attnres}), all $S = L/N$ queries within a block can be batched into a single matrix multiplication, amortizing memory access from $S$ reads to $1$.

Algorithm~\ref{alg:inference} instantiates a two-phase computation strategy exploiting this property.
\begin{itemize}[leftmargin=*]
    \item \textbf{Phase~1} computes inter-block attention for all $S$ layers simultaneously via a single batched query against the cached block representations, returning both outputs and $\operatorname{softmax}$ statistics (max and log-sum-exp). This amortizes the memory access cost, reducing reads from $S$ times to just once per block.
    \item \textbf{Phase~2} computes intra-block attention sequentially for each layer using the evolving partial sum, then merges with Phase~1 outputs through online $\operatorname{softmax}$~\citep{milakov2018online}. Because the online-$\operatorname{softmax}$ merge is elementwise, this phase naturally admits kernel fusion with surrounding operations, further reducing I/O overhead.
\end{itemize}

\begin{algorithm}[t]
    \caption{Two-phase computation for block $n$}
    \label{alg:inference}
    \KwIn{Pseudo queries $\{\bm{w}_l\}_{l \in \mathcal{B}_n}$, block representations $\{\bm{b}_0, \ldots, \bm{b}_{n-1}\}$}
    \BlankLine
    \tcc{Phase 1: Parallel inter-block attention}
    $\mathbf{Q} \gets [\bm{w}_l]_{l \in \mathcal{B}_n}$\tcp*{$[S, d]$}
    $\mathbf{K}, \mathbf{V} \gets [\bm{b}_0; \ldots; \bm{b}_{n-1}]$\tcp*{$[n, d]$}
    $\{\bm{o}_l^{(1)}, m_l^{(1)}, \ell_l^{(1)}\}_{l \in \mathcal{B}_n} \gets \textsc{AttnWithStats}(\mathbf{Q}, \mathbf{K}, \mathbf{V})$\tcp*{Return LSE}
    \;
    \tcc{Phase 2: Sequential intra-block attention + Online $\operatorname{softmax}$ merge}
    $i \gets 0$\;
    \For{$l \in \mathcal{B}_n$}{
        \eIf{$i = 0$}{
            $\bm{h}_l \gets \bm{o}_l^{(1)} / \ell_l^{(1)}$\tcp*{Inter-block only}
        }{
            $\bm{o}_l^{(2)}, m_l^{(2)}, \ell_l^{(2)} \gets \textsc{AttnWithStats}(\bm{w}_l, \bm{b}_n^i, \bm{b}_n^i)$\tcp*{Intra-block}
            $m_l \gets \max(m_l^{(1)}, m_l^{(2)})$\;
            $\bm{h}_l \gets \dfrac{e^{m_l^{(1)} - m_l} \bm{o}_l^{(1)} + e^{m_l^{(2)} - m_l} \bm{o}_l^{(2)}}{e^{m_l^{(1)} - m_l} \ell_l^{(1)} + e^{m_l^{(2)} - m_l} \ell_l^{(2)}}$\tcp*{Online softmax merge}
        }
        $i \gets i + 1$\;
        $\bm{b}_n^{i} \gets \bm{b}_n^{i-1} + f_l(\bm{h}_l)$\tcp*{Update partial sum; $\bm{b}_n^0 \coloneqq \bm{0}$}
    }
    \KwRet{$\{\bm{h}_l\}_{l \in \mathcal{B}_n}$}
\end{algorithm}

With the two-phase design, Phase~2 preserves an I/O footprint similar to that of standard residual connections, whereas the main additional cost arises from Phase~1 inter-block attention. Because these inter-block reads are amortized across all layers in a block through batching, the total per-layer memory access cost remains only $(\frac{N}{S} + 3)d$ reads and $2d$ writes (Table~\ref{tab:memory_access}). This is substantially lower than the residual-stream I/O of prior residual generalizations such as (m)HC under typical settings. In practice, Phase~1 can also partially overlap with the computation of the first layer in the block, further reducing its wall-clock impact. As a result, the end-to-end inference latency overhead is less than 2\% on typical inference workloads.

\begin{table}[h]
  \centering
  \small
  \setlength{\tabcolsep}{9pt}
  \renewcommand{\arraystretch}{1.1}
  \caption{Memory access cost per token per layer incurred by the residual mechanism under each scheme. The internal I/O of the layer function $f_l$ is excluded. For AttnRes, both Full and Block variants use the two-phase inference schedule described in Appendix~\ref{app:inference}; amortized costs are averaged over $N$ layers within a block. Typical values: $L{=}128$, $N{=}8$, $S{=}L/N{=}16$, $m{=}4$.}
  \label{tab:memory_access}
  \newlength{\iocolwidth}\settowidth{\iocolwidth}{$(8m{+}2)d{+}2m^2{+}4m$}%
  \begin{tabular}{@{}l l l c c w{c}{\iocolwidth} w{c}{\iocolwidth}@{}}
    \toprule
                                                           &                                                               & \multirow{2}{*}{Operation} & \multirow{2}{*}{Read}           & \multirow{2}{*}{Write}                   & \multicolumn{2}{c}{Total I/O}                                                \\
    \cmidrule(l){6-7}
                                                           &                                                               &                            &                                 &                                          & Symbolic                                           & Typical                 \\
    \midrule
    \multicolumn{2}{l}{Standard Residuals}                 & Residual Merge                                                & $2d$                       & $d$                             & $3d$                                     & $3d$                                                                         \\
    \midrule
    \multicolumn{2}{l}{\multirow{5}{*}{mHC ($m$ streams)}} & Compute $\bm{\alpha}_{l}$, $\bm{\beta}_{l}$, $\mathbf{A}_{l}$ & $md$                       & $m^2{+}2m$                      & \multirow{5}{*}{$(8m{+}2)d{+}2m^2{+}4m$} & \multirow{5}{*}{$34d$}                                                       \\
    \multicolumn{2}{l}{}                                   & Apply $\bm{\alpha}_{l}$                                       & $md{+}m$                   & $d$                             &                                          &                                                                              \\
    \multicolumn{2}{l}{}                                   & Apply $\bm{\beta}_{l}$                                        & $d{+}m$                    & $md$                            &                                          &                                                                              \\
    \multicolumn{2}{l}{}                                   & Apply $\mathbf{A}_{l}$                                        & $md{+}m^2$                 & $md$                            &                                          &                                                                              \\
    \multicolumn{2}{l}{}                                   & Residual Merge                                                & $2md$                      & $md$                            &                                          &                                                                              \\
    \midrule
    \multirow{4}{*}{AttnRes}                               
                                                           & \multirow{2}{*}{Full}                                         & Phase~1 (amortized)        & $(N{-}1)d$                      & $d$                                      & \multirow{2}{*}{$(S{+}N)d$}                        & \multirow{2}{*}{$24d$}  \\
                                                           &                                                               & Phase~2                    & $(S{-}1)d$                      & $d$                                      &                                                    &                         \\
    \cmidrule{2-7}
                                                           & \multirow{2}{*}{Block}                                        & Phase~1 (amortized)        & $\frac{N}{S}d$                  & $d$                                      & \multirow{2}{*}{$\left(\frac{N}{S}{+}5\right)d$}   & \multirow{2}{*}{$5.5d$} \\
                                                           &                                                               & Phase~2                    & $3d$                            & $d$                                      &                                                    &                         \\
    \bottomrule
  \end{tabular}
\end{table}

\paragraph{Memory-efficient prefilling.}
Storing block representations during prefilling requires $N \cdot T \cdot d$ elements, which incurs 15\,GB of memory for a 128K-token sequence with 8 blocks. We mitigate this by sharding these representations along the sequence dimension across $P$ tensor-parallel devices, allowing Phase~1 to execute independently on local sequence shards. The Phase~2 online-$\operatorname{softmax}$ merge then integrates into the standard TP all-reduce communication path: the output is reduce-scattered, merged locally, and reconstructed via all-gather, naturally admitting kernel fusion with operations like $\operatorname{RMSNorm}$. This reduces the per-device memory footprint to $N \cdot (T/P) \cdot d$---lowering the 128K-context example from 15\,GB to roughly 1.9\,GB per device. Combined with chunked prefill (e.g., 16K chunk size), the overhead further reduces to under 0.3\,GB per device.

\section{Experiments}
\label{sec:exp}

\paragraph{Architecture Details.}
Our architecture is identical to Kimi Linear~\citep{zhang2025kimi}, a Mixture-of-Experts (MoE) Transformer following the Moonlight~\citep{liu-2025-moonlight} / DeepSeek-V3~\citep{deepseekaiv3} design, which interleaves Kimi Delta Attention (KDA) and Multi-Head Latent Attention (MLA) layers in a 3:1 ratio, each followed by an MoE feed-forward layer.
The only modification is the addition of AttnRes to the residual connections; all other components (model depth, hidden dimensions, expert routing, and MLP structure) remain unchanged.
AttnRes introduces only one $\operatorname{RMSNorm}$ and one pseudo-query vector $\bm{w}_l \in \mathbb{R}^d$ per layer, amounting to a negligible fraction of the total parameter count. Crucially, all pseudo-query vectors must be initialized to zero. This ensures that the initial attention weights $\brickred{\alpha_{i \to l}}$ are uniform across source layers, which reduces AttnRes to an equal-weight average at the start of training and prevents training volatility, as we validated empirically.

\subsection{Scaling Laws}

\label{sec:scaling}

\begin{table}[t]
    \centering
    \small
    \setlength{\tabcolsep}{6pt}
    \caption{Baseline vs Block AttnRes ($N=8$) vs Full AttnRes vs mHC(-lite) \cite{yang2026mhclite}: Model configurations, Hyperparameters, and Validation Loss.
    }
    \label{tab:baseline_mhc_attnres_compact}
    \resizebox{\textwidth}{!}{%
        \begin{tabular}{cccccccc cccc}
            \toprule
            \multirow{2}{*}{\shortstack{\# Act.                                                                                                                                              \\Params$^\dagger$}} & \multirow{2}{*}{Tokens} & \multirow{2}{*}{$L_b$} & \multirow{2}{*}{$H$} & \multirow{2}{*}{$d_{\text{model}}$} & \multirow{2}{*}{$d_{\text{ff}}$} & \multirow{2}{*}{lr}   & \multirow{2}{*}{batch size$^\ddagger$} &
            \multicolumn{4}{c}{Val.\ Loss}                                                                                                                                                   \\
            \cmidrule(lr){9-12}
                 &                &    &    &              &     &                       &     & Baseline & \shortstack{Block AttnRes} & \shortstack{Full AttnRes} & \shortstack{mHC(-lite)} \\
            \midrule
            194M & \white{0}38.7B & 12 & 12 & \white{0}896 & 400 & $2.99 \times 10^{-3}$ & 192 & 1.931    & 1.909                      & \textbf{1.899}            & 1.906                   \\
            241M & \white{0}45.4B & 13 & 13 & \white{0}960 & 432 & $2.80 \times 10^{-3}$ & 256 & 1.895    & 1.875                      & 1.874                     & \textbf{1.869}          \\
            296M & \white{0}62.1B & 14 & 14 & 1024         & 464 & $2.50 \times 10^{-3}$ & 320 & 1.829    & 1.809                      & \textbf{1.804}            & 1.807                   \\
            436M & \white{0}87.9B & 16 & 16 & 1168         & 528 & $2.20 \times 10^{-3}$ & 384 & 1.766    & 1.746                      & \textbf{1.737}            & 1.747                   \\
            528M & 119.0B         & 17 & 17 & 1264         & 560 & $2.02 \times 10^{-3}$ & 432 & 1.719    & 1.693                      & \textbf{1.692}            & 1.694                   \\
            \bottomrule
            \multicolumn{12}{l}{\rule{0pt}{3ex}$^\dagger$ Denotes the number of activated parameters in our MoE models, excluding embeddings.}                                               \\
            \multicolumn{12}{l}{$^\ddagger$ All models were trained with a context length of 8192.}                                                                                          \\
            \multicolumn{12}{l}{$^\star$ $L_b = L/2$ denotes the number of Transformer blocks.}
        \end{tabular}}
\end{table}

\pgfdeclareplotmark{redstar}{
    \node[star,star point ratio=2.25,minimum size=5pt,
        inner sep=0pt,fill=brickred!80] {};
}
\pgfdeclareplotmark{bluestar}{
    \node[star,star point ratio=2.25,minimum size=5pt,
        inner sep=0pt,fill=midnightblue!80] {};
}
\definecolor{seagreen}{RGB}{235, 165, 85}
\pgfdeclareplotmark{greenstar}{
    \node[star,star point ratio=2.25,minimum size=5pt,
        inner sep=0pt,fill=seagreen] {};
}

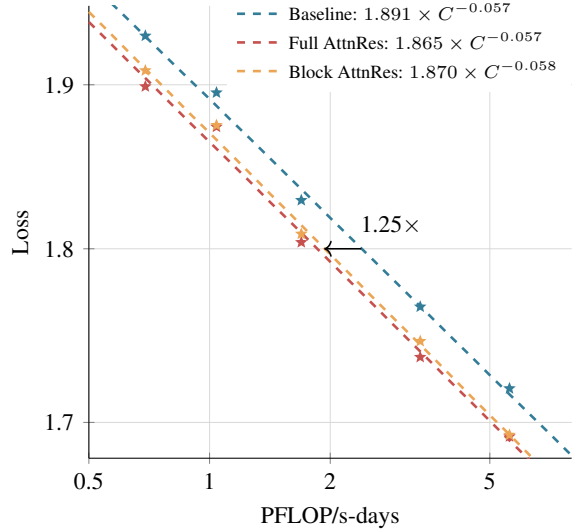
\begin{figure}[t]
    \centering
    \begin{tikzpicture}
        \begin{axis}[
                xmode=log,
                ymode=log,
                log ticks with fixed point,
                x filter/.code=\pgfmathparse{#1},
                xlabel={PFLOP/s-days},
                ylabel={Loss},
                xlabel style={font=\small},
                ylabel style={font=\small, at={(0.05,0.5)}},
                xmin=0.5,
                xmax=8,
                ymin=1.68,
                ymax=1.95,
                xtick={0.5, 1, 2, 5},
                xticklabels={0.5, 1, 2, 5},
                ytick={1.7, 1.8, 1.9},
                yticklabels={1.7, 1.8, 1.9},
                width=8cm, height=7.6cm,
                scaled y ticks=false,
                grid=major,
                tick label style={font=\small},
                axis x line=bottom,
                axis y line=left,
                axis line style={-, line width=0.4pt},
                major grid style={line width=0.15pt, gray!30},
                legend style={
                        at={(1,1.03)},
                        anchor=north east,
                        legend cell align=left,
                        font=\scriptsize,
                        draw=none,
                        row sep=-1pt,
                    },
            ]
            \addplot[midnightblue!80, dashed, line width=1.0pt, domain=0.5:8] {
                1.8908 * exp(-0.0565 * ln(x))
            };
            \addlegendentry{Baseline: $1.891 \times C^{-0.057}$}

            \addplot[brickred!80, dashed, line width=1.0pt, domain=0.5:8] {
                1.8645 * exp(-0.0572 * ln(x))
            };
            \addlegendentry{Full AttnRes: $1.865 \times C^{-0.057}$}

            \addplot[seagreen, dashed, line width=1.0pt, domain=0.5:8] {
                1.8699 * exp(-0.0578 * ln(x))
            };
            \addlegendentry{Block AttnRes: $1.870 \times C^{-0.058}$}

            \addplot[color=midnightblue!80, only marks, mark=bluestar] coordinates {
                    (0.6916, 1.931)
                    (1.0405, 1.895)
                    (1.6937, 1.829)
                    (3.3528, 1.766)
                    (5.5974, 1.719)
                };

            \addplot[color=brickred!80, only marks, mark=redstar] coordinates {
                    (0.6916, 1.899)
                    (1.0405, 1.874)
                    (1.6937, 1.804)
                    (3.3528, 1.737)
                    (5.5974, 1.692)
                };

            \addplot[color=seagreen, only marks, mark=greenstar] coordinates {
                    (0.6916, 1.909)
                    (1.0405, 1.875)
                    (1.6937, 1.809)
                    (3.3528, 1.746)
                    (5.5974, 1.693)
                };

            \draw[->, semithick, black] (axis cs:2.39, 1.80) -- (axis cs:1.93, 1.80)
            node[pos=-0.8, above, yshift=2pt, font=\small] {1.25$\times$};

        \end{axis}
    \end{tikzpicture}
    \caption{Scaling law curves for Attention Residuals. Both Full and Block AttnRes consistently outperform the baseline across all scales. Block AttnRes closely tracks Full AttnRes, recovering most of the gain at the largest scale.}
    \label{fig:scaling_law}
\end{figure}

We sweep five model sizes (Table~\ref{tab:baseline_mhc_attnres_compact}) and train three variants per size: a PreNorm baseline, Full AttnRes, and Block AttnRes with ${\approx}\,8$ blocks.
They are trained with an 8192-token context window and a cosine learning rate schedule.
Within each scaling law size group, all variants share identical hyperparameters selected under the baseline to ensure fair comparison; this setup intentionally favors the baseline and thus makes the comparison conservative. Following standard practice, we fit power-law curves of the form $\mathcal{L} = A \times C^{-\alpha}$ \cite{kaplan2020scalinglawsneurallanguage,hoffmann2022trainingcomputeoptimallargelanguage}, where $\mathcal{L}$ is validation loss and $C$ is compute measured in PFLOP/s-days.

\paragraph{Scaling Behavior.}
Fig.~\ref{fig:scaling_law} presents the fitted scaling curves. The Baseline follows $\mathcal{L} = 1.891 \times C^{-0.057}$, while Block AttnRes fits $\mathcal{L} = 1.870 \times C^{-0.058}$, and Full AttnRes fits $\mathcal{L} = 1.865 \times C^{-0.057}$. All three variants exhibit a similar slope, but AttnRes consistently achieves lower loss across the entire compute range. Based on the fitted curves, at 5.6 PFLOP/s-days, Block AttnRes reaches 1.692 versus the Baseline's 1.714, equivalent to a $1.25\times$ compute advantage. The gap between Full and Block AttnRes narrows with scale, shrinking to just 0.001 at the largest size.
We also list mHC(-lite)~\cite{yang2026mhclite} in Table~\ref{tab:baseline_mhc_attnres_compact} for reference. Full AttnRes outperforms mHC, while Block AttnRes matches it at lower memory I/O per layer: $5.5d$ versus $34d$ for mHC with $m{=}4$ streams (Table~\ref{tab:memory_access}).

\subsection{Main Results}
\paragraph{Training recipe.}
The largest models we study are based on the full Kimi Linear 48B configuration: 27 Transformer blocks (54 layers) with 8 out of 256 routed experts plus 1 shared expert, yielding 48B total and 3B activated parameters.
This model applies Block AttnRes with 6 layers per block, producing 9 blocks plus the token embedding for a total of 10 depth-wise sources.

We follow the same data and training recipe as the Kimi Linear 1.4T-token runs~\citep{zhang2025kimi}: all models are pre-trained with a 4096-token context window, the Muon optimizer~\citep{liu-2025-moonlight}, and a WSD (Warmup--Stable--Decay) learning rate schedule~\citep{hu2024minicpm}, with a global batch size of 8M tokens. 
Training of the final model proceeds in two stages: (i) a WSD pre-training phase on 1T tokens, followed by (ii) a mid-training phase on ${\approx}\,$400B high-quality tokens, following the annealing recipe of Moonlight~\citep{liu-2025-moonlight}.

After mid-training, we continue training with progressively longer sequence length of 32K tokens. Since our architecture uses hybrid KDA/MLA attention~\citep{zhang2025kimi}, where MLA operates without positional encodings (NoPE)~\cite{yang2025ropenopeagainnew}, context extension requires no modifications such as YaRN~\citep{peng2023yarn} or attention temperature rescaling.

\label{sec:main}

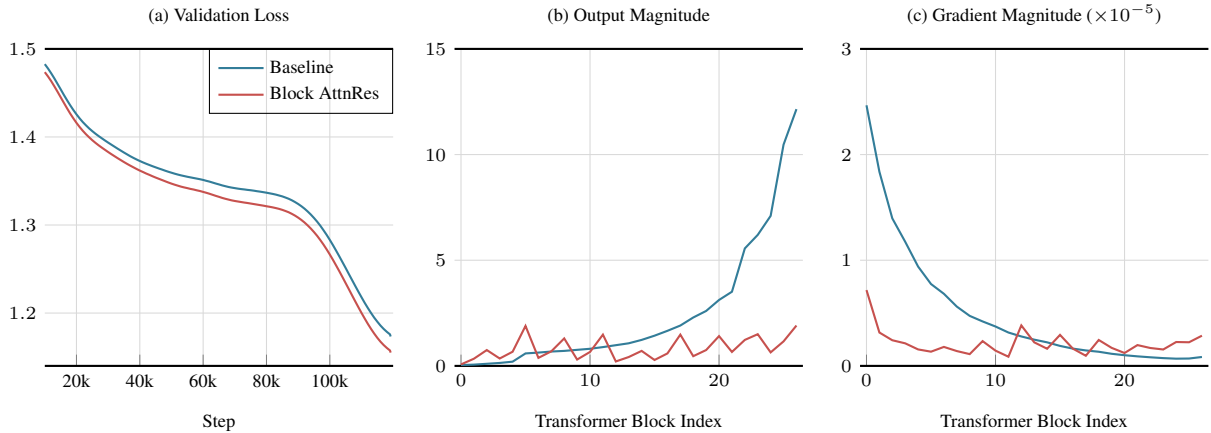
\begin{figure*}[h]
    \centering
    \small
    \pgfplotsset{
        every axis/.style={
                width=0.85\linewidth,
                height=120pt,
                scale only axis,
                grid=major,
                grid style={black!15, thin},
                axis x line*=bottom,
                x axis line style={line width=0.8pt},
                axis y line*=left,
                y axis line style={draw=none},
                every y tick/.style={draw=none},
                xlabel style={font=\scriptsize},
                ylabel=,
                title style={font=\scriptsize, at={(0.5,1)}, anchor=south},
                xticklabel style={font=\scriptsize},
                yticklabel style={font=\scriptsize},
                legend style={font=\scriptsize, fill=none, legend cell align=left},
                scaled x ticks=false,
                enlargelimits=false,
                clip=true,
            },
        baseline/.style={midnightblue!80, line width=0.8pt, mark=hexagon*, mark size=0.8pt, mark options={fill=midnightblue!80}},
        attnres/.style={brickred!80, line width=0.8pt, mark=hexagon*, mark size=0.8pt, mark options={fill=brickred!80}},
    }
    \begin{subfigure}[t]{0.33\textwidth}
        \begin{tikzpicture}[baseline=(loss.north)]
            \begin{axis}[
                    name=loss,
                    title={(a) Validation Loss},
                    xlabel={Step},
                    xmin=10000, xmax=120000,
                    ymin=1.14, ymax=1.50,
                    xtick={20000,40000,60000,80000,100000},
                    xticklabels={20k,40k,60k,80k,100k},
                    legend style={at={(1.0,1.0)}, anchor=north east},
                ]
                \addplot[baseline, mark=none] coordinates {
                        (10000,1.4827) (11000,1.4778) (12000,1.4724) (13000,1.4666) (14000,1.4604) (15000,1.4542)
                        (16000,1.4479) (17000,1.4419) (18000,1.4361) (19000,1.4307) (20000,1.4257) (21000,1.4211)
                        (22000,1.4169) (23000,1.4132) (24000,1.4097) (25000,1.4066) (26000,1.4037) (27000,1.401)
                        (28000,1.3985) (29000,1.396) (30000,1.3936) (31000,1.3913) (32000,1.389) (33000,1.3867)
                        (34000,1.3845) (35000,1.3823) (36000,1.3802) (37000,1.3781) (38000,1.3762) (39000,1.3744)
                        (40000,1.3727) (41000,1.3711) (42000,1.3696) (43000,1.3682) (44000,1.3668) (45000,1.3655)
                        (46000,1.3642) (47000,1.363) (48000,1.3617) (49000,1.3605) (50000,1.3594) (51000,1.3583)
                        (52000,1.3573) (53000,1.3564) (54000,1.3556) (55000,1.3548) (56000,1.3541) (57000,1.3534)
                        (58000,1.3527) (59000,1.352) (60000,1.3512) (61000,1.3502) (62000,1.3493) (63000,1.3482)
                        (64000,1.3471) (65000,1.3461) (66000,1.345) (67000,1.3441) (68000,1.3432) (69000,1.3425)
                        (70000,1.3419) (71000,1.3413) (72000,1.3408) (73000,1.3404) (74000,1.3399) (75000,1.3395)
                        (76000,1.339) (77000,1.3384) (78000,1.3379) (79000,1.3373) (80000,1.3367) (81000,1.336)
                        (82000,1.3353) (83000,1.3345) (84000,1.3335) (85000,1.3325) (86000,1.3312) (87000,1.3298)
                        (88000,1.3281) (89000,1.3261) (90000,1.3239) (91000,1.3214) (92000,1.3186) (93000,1.3154)
                        (94000,1.3119) (95000,1.308) (96000,1.3038) (97000,1.2992) (98000,1.2941) (99000,1.2888)
                        (100000,1.2831) (101000,1.2771) (102000,1.2708) (103000,1.2644) (104000,1.2578) (105000,1.2512)
                        (106000,1.2444) (107000,1.2377) (108000,1.231) (109000,1.2244) (110000,1.218) (111000,1.2117)
                        (112000,1.2058) (113000,1.2002) (114000,1.1949) (115000,1.1902) (116000,1.1858) (117000,1.182)
                        (118000,1.1787) (119000,1.1758) (119209,1.1734)
                    };
                \addplot[attnres, mark=none] coordinates {
                        (10000,1.4736) (11000,1.4688) (12000,1.4635) (13000,1.4576) (14000,1.4515) (15000,1.4452)
                        (16000,1.4389) (17000,1.4327) (18000,1.4267) (19000,1.4212) (20000,1.416) (21000,1.4113)
                        (22000,1.407) (23000,1.4031) (24000,1.3996) (25000,1.3963) (26000,1.3933) (27000,1.3905)
                        (28000,1.3879) (29000,1.3854) (30000,1.3829) (31000,1.3805) (32000,1.3782) (33000,1.3759)
                        (34000,1.3737) (35000,1.3715) (36000,1.3694) (37000,1.3674) (38000,1.3655) (39000,1.3636)
                        (40000,1.3618) (41000,1.3602) (42000,1.3585) (43000,1.357) (44000,1.3554) (45000,1.3539)
                        (46000,1.3525) (47000,1.3511) (48000,1.3497) (49000,1.3483) (50000,1.347) (51000,1.3458)
                        (52000,1.3447) (53000,1.3436) (54000,1.3427) (55000,1.3419) (56000,1.341) (57000,1.3403)
                        (58000,1.3394) (59000,1.3386) (60000,1.3376) (61000,1.3366) (62000,1.3355) (63000,1.3343)
                        (64000,1.3331) (65000,1.332) (66000,1.3308) (67000,1.3298) (68000,1.3289) (69000,1.328)
                        (70000,1.3273) (71000,1.3267) (72000,1.3261) (73000,1.3255) (74000,1.325) (75000,1.3244)
                        (76000,1.3238) (77000,1.3232) (78000,1.3226) (79000,1.322) (80000,1.3213) (81000,1.3207)
                        (82000,1.32) (83000,1.3192) (84000,1.3183) (85000,1.3172) (86000,1.3159) (87000,1.3145)
                        (88000,1.3127) (89000,1.3107) (90000,1.3084) (91000,1.3058) (92000,1.3029) (93000,1.2996)
                        (94000,1.296) (95000,1.292) (96000,1.2877) (97000,1.2829) (98000,1.2778) (99000,1.2723)
                        (100000,1.2665) (101000,1.2604) (102000,1.254) (103000,1.2475) (104000,1.2408) (105000,1.234)
                        (106000,1.2272) (107000,1.2203) (108000,1.2135) (109000,1.2068) (110000,1.2002) (111000,1.1939)
                        (112000,1.1878) (113000,1.1821) (114000,1.1768) (115000,1.1719) (116000,1.1675) (117000,1.1637)
                        (118000,1.1604) (119000,1.1575) (119209,1.1551)
                    };
                \legend{Baseline, Block AttnRes}
            \end{axis}
            \draw[black, line width=0.8pt] (loss.north west) -- (loss.north east);
        \end{tikzpicture}
        \phantomcaption\label{fig:comparison-loss}
    \end{subfigure}%
    \hfill%
    \begin{subfigure}[t]{0.33\textwidth}
        \begin{tikzpicture}[baseline=(rms.north)]
            \begin{axis}[
                    name=rms,
                    title={(b) Output Magnitude},
                    xlabel={Transformer Block Index},
                    xmin=-0.5, xmax=26.5,
                    ymin=0, ymax=15,
                    ytick={0,5,10,15},
                ]
                \addplot[baseline, mark=none] coordinates {
                        (0,0.04) (1,0.06) (2,0.10) (3,0.14) (4,0.20) (5,0.59) (6,0.63) (7,0.68) (8,0.71)
                        (9,0.76) (10,0.81) (11,0.89) (12,0.98) (13,1.07) (14,1.23) (15,1.43) (16,1.66) (17,1.91)
                        (18,2.29) (19,2.60) (20,3.12) (21,3.51) (22,5.56) (23,6.20) (24,7.10) (25,10.47) (26,12.15)
                    };
                \addplot[attnres, mark=none] coordinates {
                        (0,0.07) (1,0.34) (2,0.75) (3,0.35) (4,0.67) (5,1.89) (6,0.38) (7,0.69) (8,1.30)
                        (9,0.30) (10,0.66) (11,1.48) (12,0.21) (13,0.42) (14,0.71) (15,0.28) (16,0.59) (17,1.48)
                        (18,0.46) (19,0.75) (20,1.41) (21,0.66) (22,1.23) (23,1.50) (24,0.64) (25,1.16) (26,1.91)
                    };
            \end{axis}
            \draw[black, line width=0.8pt] (rms.north west) -- (rms.north east);
        \end{tikzpicture}
        \phantomcaption\label{fig:comparison-rms}
    \end{subfigure}%
    \hfill%
    \begin{subfigure}[t]{0.33\textwidth}
        \begin{tikzpicture}[baseline=(grad.north)]
            \begin{axis}[
                    name=grad,
                    title={(c) Gradient Magnitude ($\times 10^{-5}$)},
                    xlabel={Transformer Block Index},
                    xmin=-0.5, xmax=26.5,
                    ymin=0, ymax=3,
                    ytick={0,1,2,3},
                ]
                \addplot[baseline, mark=none] coordinates {
                        (0,2.4663) (1,1.8375) (2,1.3963) (3,1.1747) (4,0.9401) (5,0.7749) (6,0.6823)
                        (7,0.5613) (8,0.4734) (9,0.4208) (10,0.3728) (11,0.3159) (12,0.2787) (13,0.2485)
                        (14,0.2230) (15,0.1892) (16,0.1636) (17,0.1464) (18,0.1340) (19,0.1146) (20,0.1011)
                        (21,0.0904) (22,0.0815) (23,0.0737) (24,0.0684) (25,0.0698) (26,0.0844)
                    };
                \addplot[attnres, mark=none] coordinates {
                        (0,0.7177) (1,0.3159) (2,0.2428) (3,0.2136) (4,0.1559) (5,0.1346) (6,0.1797)
                        (7,0.1404) (8,0.1116) (9,0.2345) (10,0.1444) (11,0.0869) (12,0.3832) (13,0.2231)
                        (14,0.1624) (15,0.2936) (16,0.1630) (17,0.0959) (18,0.2454) (19,0.1708) (20,0.1221)
                        (21,0.1970) (22,0.1694) (23,0.1553) (24,0.2260) (25,0.2237) (26,0.2859)
                    };
            \end{axis}
            \draw[black, line width=0.8pt] (grad.north west) -- (grad.north east);
        \end{tikzpicture}
        \phantomcaption\label{fig:comparison-grad}
    \end{subfigure}

    \caption{Training dynamics of Baseline and Block AttnRes. \textbf{(\subref{fig:comparison-loss})} Validation loss during training. \textbf{(\subref{fig:comparison-rms})} Each transformer block's output magnitude at the end of training. \textbf{(\subref{fig:comparison-grad})} Each transformer block's gradient magnitude.
    }
    \label{fig:comparison}
\end{figure*}

\paragraph{Training dynamics.}
We compare the training dynamics of our final Baseline and Block AttnRes models over 1T tokens in Fig.~\ref{fig:comparison}.
\begin{itemize}[leftmargin=*]
    \item \textbf{Validation loss:} AttnRes achieves consistently lower validation loss throughout training, with the gap widening during the decay phase and resulting in a notably lower final loss.
    \item \textbf{Output magnitude:} The Baseline suffers from the PreNorm dilution problem~\citep{xiong2020layer,li2026siamesenorm}: as hidden-state magnitudes grow monotonically with depth, deeper layers are compelled to learn increasingly large outputs from fixed-scale normalized inputs to remain influential. Block AttnRes confines this growth within each block, as selective aggregation at block boundaries resets the accumulation, yielding a bounded periodic pattern.
    \item \textbf{Gradient magnitude:} With all residual weights fixed to 1, the Baseline provides no means of regulating gradient flow across depth, leading to disproportionately large gradients in the earliest layers. The learnable $\operatorname{softmax}$ weights in Block AttnRes (Fig.~\ref{fig:attn-res-weights}) introduce competition among sources for probability mass, resulting in a substantially more uniform gradient distribution.
\end{itemize}

\begin{table}[!ht]
    \centering
    \small
    \renewcommand{\arraystretch}{1.1}
    \caption{Performance comparison of AttnRes with the baseline, both after the same pre-training recipe. Best per-row results are \textbf{bolded}.}
    \setlength{\tabcolsep}{12pt}
    \label{tab:benchmarkresult}
    \begin{tabular}{@{}r l c c}
        \toprule
         &               & Baseline      & AttnRes       \\
        \midrule
        \multirow{7}{*}{\textit{General}}
         & MMLU          & 73.5          & \textbf{74.6} \\
         & MMLU-Pro      & \textbf{52.2} & \textbf{52.2} \\
         & GPQA-Diamond  & 36.9          & \textbf{44.4} \\
         & BBH           & 76.3          & \textbf{78.0} \\
         & ARC-Challenge & 64.6          & \textbf{65.7} \\
         & HellaSwag     & 83.2          & \textbf{83.4} \\
         & TriviaQA      & 69.9          & \textbf{71.8} \\
        \midrule
        \multirow{6}{*}{\textit{Math \& Code}}
         & GSM8K         & 81.7          & \textbf{82.4} \\
         & MGSM          & 64.9          & \textbf{66.1} \\
         & Math          & 53.5          & \textbf{57.1} \\
         & CMath         & 84.7          & \textbf{85.1} \\
         & HumanEval     & 59.1          & \textbf{62.2} \\
         & MBPP          & 72.0          & \textbf{73.9} \\
        \midrule
        \multirow{2}{*}{\textit{Chinese}}
         & CMMLU         & 82.0          & \textbf{82.9} \\
         & C-Eval        & 79.6          & \textbf{82.5} \\
        \bottomrule
    \end{tabular}
    \label{tab:pretrain}
\end{table}

\paragraph{Downstream performance.}
Following the evaluation protocol of Kimi Linear~\cite{zhang2025kimi}, we assess both models across three areas (Table~\ref{tab:pretrain}):
\begin{itemize}[leftmargin=*]
    \item \textbf{Language understanding and reasoning}: MMLU~\cite{hendrycks-2021-mmlu}, MMLU-Pro Hard~\cite{wang2024mmlu}, GPQA-Diamond~\cite{rein2024gpqa}, BBH~\cite{suzgun2022challenging}, ARC-Challenge~\cite{clark-2018-arc}, HellaSwag~\cite{zellers-2019-hellaswag}, and TriviaQA~\cite{joshi2017triviaqa}.
    \item \textbf{Reasoning (Code and Math)}: GSM8K~\cite{gsm8k}, MGSM~\cite{mgsm}, Math~\cite{minervamath}, CMath~\cite{cmathbmk}, HumanEval~\cite{chen2021evaluatinglargelanguagemodels}, and MBPP~\cite{austin2021programsynthesislargelanguage}.
    \item \textbf{Chinese language understanding}: CMMLU~\cite{li-etal-2024-cmmlu} and C-Eval~\cite{huang2023c}.
\end{itemize}
As shown in Table~\ref{tab:pretrain}, Block AttnRes matches or outperforms the baseline on all benchmarks. The improvements are particularly pronounced on multi-step reasoning tasks such as GPQA-Diamond (+7.5) and Minerva Math (+3.6), as well as code generation such as HumanEval (+3.1), while knowledge-oriented benchmarks such as MMLU (+1.1) and TriviaQA (+1.9) also show solid gains. This pattern is consistent with the hypothesis that improved depth-wise information flow benefits compositional tasks, where later layers can selectively retrieve and build upon earlier representations.

\subsection{Ablation Study}
\pgfdeclareplotmark{hexagon*}{
    \pgfpathmoveto{\pgfqpointpolar{0}{1.1\pgfplotmarksize}}
    \pgfpathlineto{\pgfqpointpolar{60}{1.1\pgfplotmarksize}}
    \pgfpathlineto{\pgfqpointpolar{120}{1.1\pgfplotmarksize}}
    \pgfpathlineto{\pgfqpointpolar{180}{1.1\pgfplotmarksize}}
    \pgfpathlineto{\pgfqpointpolar{240}{1.1\pgfplotmarksize}}
    \pgfpathlineto{\pgfqpointpolar{300}{1.1\pgfplotmarksize}}
    \pgfpathclose
    \pgfusepathqfillstroke
}
\begin{figure}[t!]
    \centering
    \adjustbox{valign=t}{\begin{minipage}{0.45\textwidth}
            \centering
            \small
            \setlength{\tabcolsep}{10pt}
            \renewcommand{\arraystretch}{1.18}
            \captionof{table}{Ablation on key components of AttnRes (16-layer model).}
            \label{tab:ablation}
            \begin{tabular}{llc}
                \toprule
                                                                                  & Variant                                   & Loss  \\
                \midrule
                \multicolumn{2}{l}{Baseline (PreNorm)}                            & 1.766                                             \\
                \midrule
                \multicolumn{2}{l}{DenseFormer \cite{pagliardini2024denseformer}} & 1.767                                             \\
                \multicolumn{2}{l}{mHC \cite{xie2026mhc}}                         & 1.747                                             \\
                \midrule
                AttnRes                                                           & Full                                      & 1.737 \\
                                                                                  & \quad \emph{w/ input-dependent query}     & 1.731 \\
                                                                                  & \quad \emph{w/ input-independent mixing}  & 1.749 \\
                                                                                  & \quad \emph{w/ $\operatorname{sigmoid}$}  & 1.741 \\
                                                                                  & \quad \emph{w/o $\operatorname{RMSNorm}$} & 1.743 \\
                                                                                  & SWA ($W = 1 + 8$)                         & 1.764 \\
                                                                                  & Block ($S=4$)                             & 1.746 \\
                                                                                  & \quad \emph{w/ multihead ($H=16$)}        & 1.752 \\
                                                                                  & \quad \emph{w/o $\operatorname{RMSNorm}$} & 1.750 \\
                \bottomrule
            \end{tabular}
        \end{minipage}}
    \hfill
    \adjustbox{valign=t}{\begin{minipage}{0.5\textwidth}
            \centering
            \small
            \begin{tikzpicture}
                \begin{axis}[
                        xlabel={Block size ($S$)},
                        ylabel={Validation loss},
                        xlabel style={font=\small},
                        ylabel style={font=\small},
                        separate axis lines,
                        every outer x axis line/.append style={line width=0.8pt},
                        every outer y axis line/.append style={draw=none},
                        every y tick/.style={draw=none},
                        every x tick/.style={draw=none},
                        xticklabel style={font=\small},
                        yticklabel style={font=\small, /pgf/number format/fixed, /pgf/number format/precision=3, /pgf/number format/zerofill},
                        xtick={1, 2, 3, 4, 5},
                        xticklabels={32, 16, 8, 4, 2},
                        xmin=0.5,
                        xmax=5.5,
                        ymin=1.734,
                        ymax=1.770,
                        ytick={1.735, 1.740, 1.745, 1.750, 1.755, 1.760, 1.765, 1.770},
                        scaled y ticks=false,
                        width=\textwidth,
                        height=7cm,
                        grid=major,
                        grid style={gray!20, thin},
                        enlargelimits=false,
                        legend style={
                                at={(1.,1.)},
                                anchor=north east,
                                legend cell align=left,
                                font=\scriptsize,
                                draw=gray!50,
                                fill=white,
                                fill opacity=0.9,
                                text opacity=1,
                            },
                    ]

                    \addplot[gray!60, dashed, line width=0.8pt, forget plot] coordinates {(0.5, 1.766) (5.5, 1.766)};
                    \addlegendimage{gray!60, dashed, line width=0.8pt}
                    \addlegendentry{Baseline}

                    \addplot[brickred!80, dashed, line width=0.8pt, forget plot] coordinates {(0.5, 1.737) (5.5, 1.737)};
                    \addlegendimage{brickred!80, dashed, line width=0.8pt}
                    \addlegendentry{Full AttnRes}

                    \addplot[brickred!80, line width=0.8pt, mark=hexagon*, mark size=1.2pt, mark options={fill=brickred!80}] coordinates {
                            (1, 1.757)
                            (2, 1.753)
                            (3, 1.748)
                            (4, 1.746)
                            (5, 1.746)
                        };
                    \addlegendentry{Block AttnRes}

                    \node[font=\tiny, above=3pt] at (axis cs:1, 1.757) {1.757};
                    \node[font=\tiny, above=3pt] at (axis cs:2, 1.753) {1.753};
                    \node[font=\tiny, above=3pt] at (axis cs:3, 1.748) {1.748};
                    \node[font=\tiny, above=3pt] at (axis cs:4, 1.746) {1.746};
                    \node[font=\tiny, above=3pt] at (axis cs:5, 1.746) {1.746};

                    \node[font=\tiny, black, above=2pt] at (axis cs:3, 1.766) {Baseline (1.766)};

                    \node[font=\tiny, black, below=2pt,text=brickred!80] at (axis cs:3, 1.737) {Full AttnRes i.e. S=1 (1.737)};

                \end{axis}
            \end{tikzpicture}
            \captionof{figure}{Effect of block size on validation loss (16-layer model).}
            \label{fig:blocksize}
        \end{minipage}}
\end{figure}
We conduct ablation studies on the 16-head model from Table~\ref{tab:baseline_mhc_attnres_compact} to validate key design choices in AttnRes (Table~\ref{tab:ablation}). All models share identical hyperparameters and compute budget.

\paragraph{Comparison with prior methods.}
We compare AttnRes against the PreNorm baseline (loss 1.766) and two representative methods that generalize residual connections. DenseFormer~\cite{pagliardini2024denseformer} grants each layer access to all previous outputs but combines them with fixed, input-independent scalar coefficients; it shows no gain over the baseline (1.767), highlighting the importance of input-dependent weighting. mHC~\cite{xie2026mhc} introduces input dependence through $m$ parallel streams with learned mixing matrices, improving to 1.747. AttnRes takes this further with explicit content-dependent selection via $\operatorname{softmax}$ attention: Full AttnRes achieves 1.737 and Block AttnRes 1.746, outperforming both methods with only a single query vector per layer.

\paragraph{Cross-layer access.}
We compare three granularities of cross-layer access. Full AttnRes follows directly from the time--depth duality (\S~\ref{sec:attnres}), applying attention over all previous layers, and achieves the lowest loss (1.737). A simple way to reduce its memory cost is sliding-window aggregation (SWA), which retains only the most recent $W{=}8$ layer outputs plus the token embedding; it improves over baseline (1.764) but falls well short of both Full and Block AttnRes, suggesting that selectively accessing distant layers matters more than attending to many nearby ones.

Block AttnRes offers a better trade-off: with block size $S{=}4$ it reaches 1.746 while keeping memory overhead constant per layer. Fig.~\ref{fig:blocksize} sweeps $S$ across the full spectrum from $S{=}1$ (i.e.\ Full AttnRes) to increasingly coarse groupings. Loss degrades gracefully as $S$ grows, with $S{=}2, 4, 8$ all landing near 1.746 while larger blocks ($S{=}16, 32$) move toward baseline. In practice, we fix the number of blocks to ${\approx}\,8$ for infrastructure efficiency (\S~\ref{sec:infra}). As future hardware alleviates memory capacity constraints, adopting finer-grained block sizes or Full AttnRes represents a natural pathway to further improve performance.

\paragraph{Component design.}
We further ablate individual components of the attention mechanism:
\begin{itemize}[leftmargin=*]
    \item \textbf{Input-dependent query.} A natural extension is to make the query input-dependent by projecting it from the current hidden state. This further lowers loss to 1.731, but introduces a $d \times d$ projection per layer and requires sequential memory access during decoding, so we default to the learned query.
    \item \textbf{Input-independent mixing.} We removed the query and key and replaced them with learnable, input-independent scalars to weigh previous layers, which hurts performance (1.749 vs.\ 1.737).
    \item \textbf{$\operatorname{softmax}$ vs.\ $\operatorname{sigmoid}$.} Replacing $\operatorname{softmax}$ with $\operatorname{sigmoid}$ degrades performance (1.741). We attribute this to $\operatorname{softmax}$'s competitive normalization, which forces sharper selection among sources.
    \item \textbf{Multihead attention.} We test per-head depth aggregation ($H{=}16$) on Block AttnRes, allowing different channel groups to attend to different source layers. This hurts performance (1.752 vs.\ 1.746), indicating that the optimal depth-wise mixture is largely uniform across channels: when a layer's output is relevant, it is relevant as a whole.
    \item \textbf{$\operatorname{RMSNorm}$ on keys.} Removing $\operatorname{RMSNorm}$ degrades both Full AttnRes (1.743) and Block AttnRes (1.750). For Full AttnRes, it prevents individual layers with naturally larger outputs from dominating the $\operatorname{softmax}$. This becomes even more critical for Block AttnRes, as block-level representations accumulate over more layers and can develop large magnitude differences; $\operatorname{RMSNorm}$ prevents these from biasing the attention weights.
\end{itemize}

\subsection{Analysis}

\subsubsection{Optimal Architecture}
\label{sec:optimal_arch}

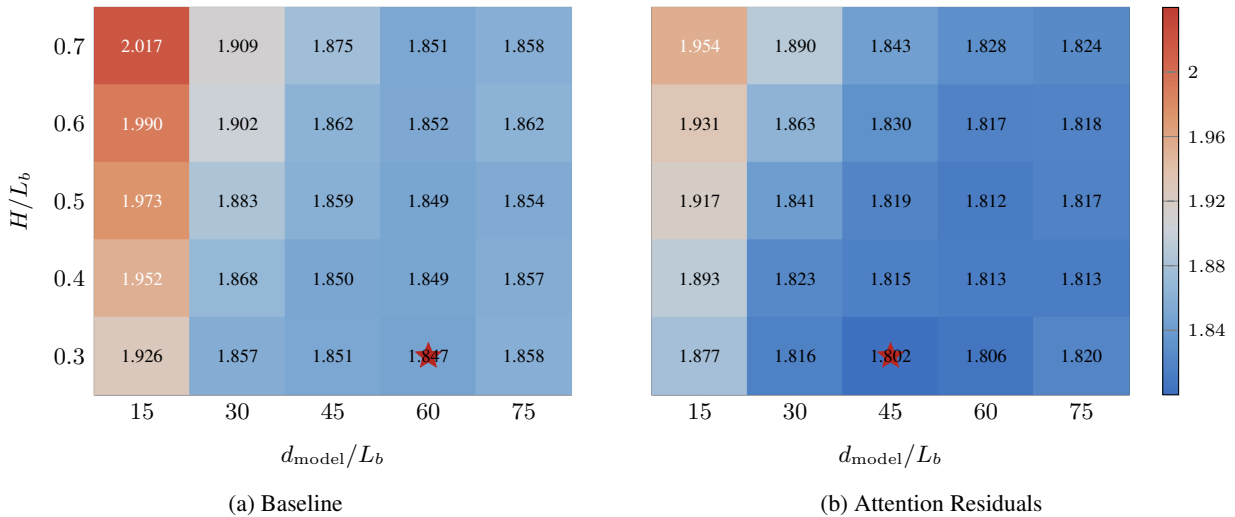
\begin{figure}[t]
  \centering
  \begin{subfigure}[b]{0.48\textwidth}
    \centering
    \begin{tikzpicture}
      \begin{axis}[
          width=\textwidth,
          height=0.85\textwidth,
          xlabel={$d_{\mathrm{model}}/L_b$},
          ylabel={$H/L_b$},
          xtick={15,30,45,60,75},
          ytick={0.3,0.4,0.5,0.6,0.7},
          xticklabel style={font=\small},
          yticklabel style={font=\small},
          xlabel style={font=\small},
          ylabel style={font=\small, at={(0.04,0.5)}},
          colormap={softcoolwarm}{
              rgb255(0)=(60,110,190)
              rgb255(1)=(100,150,205)
              rgb255(2)=(150,185,215)
              rgb255(3)=(205,210,215)
              rgb255(4)=(225,195,175)
              rgb255(5)=(220,150,110)
              rgb255(6)=(210,100,75)
              rgb255(7)=(190,60,50)
            },
          title style={font=\small},
          point meta min=1.80,
          point meta max=2.04,
          enlargelimits=false,
        ]
        \addplot[matrix plot*, mesh/cols=5, point meta=explicit] coordinates {
            (15,0.7) [2.017] (30,0.7) [1.909] (45,0.7) [1.875] (60,0.7) [1.851] (75,0.7) [1.858]
            (15,0.6) [1.990] (30,0.6) [1.902] (45,0.6) [1.862] (60,0.6) [1.852] (75,0.6) [1.862]
            (15,0.5) [1.973] (30,0.5) [1.883] (45,0.5) [1.859] (60,0.5) [1.849] (75,0.5) [1.854]
            (15,0.4) [1.952] (30,0.4) [1.868] (45,0.4) [1.850] (60,0.4) [1.849] (75,0.4) [1.857]
            (15,0.3) [1.926] (30,0.3) [1.857] (45,0.3) [1.851] (60,0.3) [1.847] (75,0.3) [1.858]
          };

        \node[font=\scriptsize, text=white] at (axis cs:15,0.7) {2.017};
        \node[font=\scriptsize, text=black] at (axis cs:30,0.7) {1.909};
        \node[font=\scriptsize, text=black] at (axis cs:45,0.7) {1.875};
        \node[font=\scriptsize, text=black] at (axis cs:60,0.7) {1.851};
        \node[font=\scriptsize, text=black] at (axis cs:75,0.7) {1.858};
        \node[font=\scriptsize, text=white] at (axis cs:15,0.6) {1.990};
        \node[font=\scriptsize, text=black] at (axis cs:30,0.6) {1.902};
        \node[font=\scriptsize, text=black] at (axis cs:45,0.6) {1.862};
        \node[font=\scriptsize, text=black] at (axis cs:60,0.6) {1.852};
        \node[font=\scriptsize, text=black] at (axis cs:75,0.6) {1.862};
        \node[font=\scriptsize, text=white] at (axis cs:15,0.5) {1.973};
        \node[font=\scriptsize, text=black] at (axis cs:30,0.5) {1.883};
        \node[font=\scriptsize, text=black] at (axis cs:45,0.5) {1.859};
        \node[font=\scriptsize, text=black] at (axis cs:60,0.5) {1.849};
        \node[font=\scriptsize, text=black] at (axis cs:75,0.5) {1.854};
        \node[font=\scriptsize, text=white] at (axis cs:15,0.4) {1.952};
        \node[font=\scriptsize, text=black] at (axis cs:30,0.4) {1.868};
        \node[font=\scriptsize, text=black] at (axis cs:45,0.4) {1.850};
        \node[font=\scriptsize, text=black] at (axis cs:60,0.4) {1.849};
        \node[font=\scriptsize, text=black] at (axis cs:75,0.4) {1.857};
        \node[font=\scriptsize, text=black] at (axis cs:15,0.3) {1.926};
        \node[font=\scriptsize, text=black] at (axis cs:30,0.3) {1.857};
        \node[font=\scriptsize, text=black] at (axis cs:45,0.3) {1.851};
        \node[font=\scriptsize, text=black] at (axis cs:75,0.3) {1.858};

        \node[star, star point ratio=2.25, minimum size=10pt,
          inner sep=0pt, draw=brickred, solid, fill=brickred]
        at (axis cs:60,0.3) {};
        \node[font=\scriptsize, text=black] at (axis cs:60,0.3) {1.847};
      \end{axis}
    \end{tikzpicture}
    \caption{Baseline}
    \label{fig:arch_sweep_baseline}
  \end{subfigure}
  \hfill
  \begin{subfigure}[b]{0.48\textwidth}
    \centering
    \begin{tikzpicture}
      \begin{axis}[
          width=\textwidth,
          height=0.85\textwidth,
          xlabel={$d_{\mathrm{model}}/L_b$},
          xtick={15,30,45,60,75},
          xticklabel style={font=\small},
          yticklabel style={font=\small},
          yticklabels={,,},
          xlabel style={font=\small},
          ylabel style={font=\small, at={(0.04,0.5)}},
          colormap={softcoolwarm}{
              rgb255(0)=(60,110,190)
              rgb255(1)=(100,150,205)
              rgb255(2)=(150,185,215)
              rgb255(3)=(205,210,215)
              rgb255(4)=(225,195,175)
              rgb255(5)=(220,150,110)
              rgb255(6)=(210,100,75)
              rgb255(7)=(190,60,50)
            },
          colorbar,
          colorbar style={
              width=6pt,
              xshift=4pt,
              ytick={1.84, 1.88, 1.92, 1.96, 2.00},
              yticklabel style={font=\scriptsize},
            },
          title style={font=\small},
          point meta min=1.80,
          point meta max=2.04,
          enlargelimits=false,
        ]
        \addplot[matrix plot*, mesh/cols=5, point meta=explicit] coordinates {
            (15,0.7) [1.954] (30,0.7) [1.890] (45,0.7) [1.843] (60,0.7) [1.828] (75,0.7) [1.824]
            (15,0.6) [1.931] (30,0.6) [1.863] (45,0.6) [1.830] (60,0.6) [1.817] (75,0.6) [1.818]
            (15,0.5) [1.917] (30,0.5) [1.841] (45,0.5) [1.819] (60,0.5) [1.812] (75,0.5) [1.817]
            (15,0.4) [1.893] (30,0.4) [1.823] (45,0.4) [1.815] (60,0.4) [1.813] (75,0.4) [1.813]
            (15,0.3) [1.877] (30,0.3) [1.816] (45,0.3) [1.802] (60,0.3) [1.806] (75,0.3) [1.820]
          };

        \node[font=\scriptsize, text=white] at (axis cs:15,0.7) {1.954};
        \node[font=\scriptsize, text=black] at (axis cs:30,0.7) {1.890};
        \node[font=\scriptsize, text=black] at (axis cs:45,0.7) {1.843};
        \node[font=\scriptsize, text=black] at (axis cs:60,0.7) {1.828};
        \node[font=\scriptsize, text=black] at (axis cs:75,0.7) {1.824};
        \node[font=\scriptsize, text=black] at (axis cs:15,0.6) {1.931};
        \node[font=\scriptsize, text=black] at (axis cs:30,0.6) {1.863};
        \node[font=\scriptsize, text=black] at (axis cs:45,0.6) {1.830};
        \node[font=\scriptsize, text=black] at (axis cs:60,0.6) {1.817};
        \node[font=\scriptsize, text=black] at (axis cs:75,0.6) {1.818};
        \node[font=\scriptsize, text=black] at (axis cs:15,0.5) {1.917};
        \node[font=\scriptsize, text=black] at (axis cs:30,0.5) {1.841};
        \node[font=\scriptsize, text=black] at (axis cs:45,0.5) {1.819};
        \node[font=\scriptsize, text=black] at (axis cs:60,0.5) {1.812};
        \node[font=\scriptsize, text=black] at (axis cs:75,0.5) {1.817};
        \node[font=\scriptsize, text=black] at (axis cs:15,0.4) {1.893};
        \node[font=\scriptsize, text=black] at (axis cs:30,0.4) {1.823};
        \node[font=\scriptsize, text=black] at (axis cs:45,0.4) {1.815};
        \node[font=\scriptsize, text=black] at (axis cs:60,0.4) {1.813};
        \node[font=\scriptsize, text=black] at (axis cs:75,0.4) {1.813};
        \node[font=\scriptsize, text=black] at (axis cs:15,0.3) {1.877};
        \node[font=\scriptsize, text=black] at (axis cs:30,0.3) {1.816};
        \node[font=\scriptsize, text=black] at (axis cs:75,0.3) {1.820};
        \node[font=\scriptsize, text=black] at (axis cs:60,0.3) {1.806};

        \node[star, star point ratio=2.25, minimum size=10pt,
          inner sep=0pt, draw=brickred, solid, fill=brickred]
        at (axis cs:45,0.3) {};
        \node[font=\scriptsize, text=black] at (axis cs:45,0.3) {1.802};

      \end{axis}
    \end{tikzpicture}
    \caption{Attention Residuals}
    \label{fig:arch_sweep_attnres}
  \end{subfigure}
  \caption{
    Architecture sweep under fixed compute ($\approx 6.5\times 10^{19}$ FLOPs, $\approx 2.3\times 10^{8}$ active parameters).
    Each cell reports validation loss for a $(d_{\mathrm{model}}/L_b,\; H/L_b)$ configuration, where $L_b = L/2$ is the number of Transformer blocks; the star marks the optimum.
  }
  \label{fig:arch_sweep}
\end{figure}

To understand how AttnRes reshapes optimal architectural scaling, we perform a controlled capacity reallocation study under a fixed compute and parameter budget. Our central question is whether AttnRes alters the preferred depth–width–attention trade-off, and in particular, given its potential strength on the depth dimension, whether it favors deeper models compared to conventional Transformer design heuristics. To isolate structural factors directly coupled to depth, we fix the per-expert MLP expansion ratio based on internal empirical observations ($d_{\mathrm{ff}}/d_{\mathrm{model}} \approx 0.45$). We further fix total training compute (FLOPs $\approx 6.5\times10^{19}$) and active parameters ($\approx 2.3\times10^8$), ensuring that any performance variation arises purely from architectural reallocation rather than overall capacity differences. Under this constrained budget, we enumerate 25 configurations on a $5\times5$ grid over
$d_{\mathrm{model}}/L_b \in \{15, 30, 45, 60, 75\}$ and
$H/L_b \in \{0.3, 0.4, 0.5, 0.6, 0.7\}$,
where $L_b = L/2$ is the number of Transformer blocks and $H$ the number of attention heads.
The results are shown in Fig.~\ref{fig:arch_sweep}.

Both heatmaps exhibit a shared pattern: loss decreases with growing $d_{\mathrm{model}}/L_b$ and shrinking $H/L_b$, and both methods reach their optima at $H/L_b \approx 0.3$.
Despite this shared trend, AttnRes achieves a lower loss than the baseline in each of the 25 configurations, by $0.019$--$0.063$.
The most apparent difference lies in the location of the optimum: the baseline achieves its lowest loss at $d_{\mathrm{model}}/L_b \approx 60$ ($1.847$), whereas AttnRes shifts it to $d_{\mathrm{model}}/L_b \approx 45$ ($1.802$).
Under a fixed parameter budget, a lower $d_{\mathrm{model}}/L_b$ corresponds to a deeper, narrower network, suggesting that AttnRes can exploit additional depth more effectively.
We note that this preference for depth does not directly translate to a deployment recommendation, as deeper models generally incur higher inference latency due to their sequential computation~\citep{pope-2022-efficiently}.
Rather, this sweep serves as a diagnostic that reveals where AttnRes benefits most, and this depth preference can be factored into the architecture selection alongside inference cost.

\subsubsection{Analyzing Learned AttnRes Patterns}

\input{figures/attn_res_weights}

\label{sec:analysis}
 We visualize the learned weights $\brickred{\alpha_{i \to l}}$ in Fig.~\ref{fig:attn-res-weights} for the 16-head model (from Table~\ref{tab:baseline_mhc_attnres_compact}) with both full and block ($N{=}8$) AttnRes.
Each heatmap shows how the $l$th attention or MLP layer (rows) allocates its attention over previous sources (columns), with pre-attention and pre-MLP layers shown separately.
We highlight three key observations:
\begin{itemize}[leftmargin=*]
    \item \textbf{Preserved locality.} Each layer attends most strongly to its immediate predecessor, yet selective off-diagonal concentrations emerge (e.g., layer 4 attending to early sources, layers 15--16 reaching back under the block setting), indicating learned skip connections beyond the standard residual path.
    \item \textbf{Layer specialization.} The embedding $\bm{h}_1$ retains non-trivial weight throughout, especially in pre-attention layers. Pre-MLP inputs show sharper diagonal reliance on recent representations, while pre-attention inputs maintain broader receptive fields, consistent with attention routing information across layers and MLPs operating locally.
    \item \textbf{Block AttnRes preserves structure.} Diagonal dominance, embedding persistence, and layer specialization all transfer from the full to the block variant, suggesting that block-wise compression acts as implicit regularization while preserving the essential information pathways.
\end{itemize}

\section{Discussions}

\subsection{Sequence-Depth Duality}
\label{sec:depth-ttt}

Residual connections propagate information over depth via a fixed recurrence $\bm{h}_{l} = \bm{h}_{l-1} + f_{l-1}(\bm{h}_{l-1})$, much as RNNs propagate information over time.
Test-Time Training (TTT)~\cite{sun-2024-learning} formalizes the sequence side of this analogy (cf.\ Fast Weight Programmers~\cite{schmidhuber1992learning,munkhdalai-2019-metalearned}), casting each recurrent step as gradient descent on a self-supervised loss:
\begin{equation}
    \mathbf{W}_t = \mathbf{W}_{t-1} - \eta\,\nabla\ell(\mathbf{W}_{t-1};\, \bm{x}_t),
\end{equation}
where a slow network parameterizes $\ell$ and the state $\mathbf{W}$ is updated once per token.
When $f$ is linear, this reduces to vanilla linear attention $\mathbf{S}_{t} = \mathbf{S}_{t-1} + \bm{k}_t\bm{v}_t^\top$.
The standard residual exhibits the same additive form along depth, with $\bm{h}_l$ serving as the state and each layer $f_l$ acting as one ``gradient step.''

As noted by~\cite{chen2026postlayernorm}, this duality extends to richer variants (Table~\ref{tab:residual_compare}).
Data-dependent gates on the sequence side~\cite{sun-2023-retnet,yang-etal-2024-gla} correspond to Highway networks~\cite{srivastava2015highway} on the depth side;
the delta rule~\cite{schlag-2021-deltanet,yang-2025-gdn,zhang2025kimi} corresponds to DDL~\cite{zhang2026ddl};
and MRLA~\cite{fang2023cross} mirrors GLA's~\cite{yang-etal-2024-gla} gated linear attention.
These methods all refine the recurrent update while remaining within the recurrence paradigm.
AttnRes goes a step further and replaces depth-wise recurrence with direct cross-layer attention, just as Transformers replaced temporal recurrence with self-attention.
Since the number of layers in current architectures remains well within the practical regime of $\operatorname{softmax}$ attention, we adopt vanilla depth-wise attention. Incorporating more expressive yet memory-efficient (e.g. linear-complexity) alternatives is a natural direction for future work.

\begin{table*}[t]
  \centering
  \caption{
    Comparison of residual update mechanisms. \emph{Weight}: whether the mixing coefficients are architecture-fixed, learned-static (fixed after training), or input-dependent (dynamic). \emph{Source}: which earlier representations layer $l$ can access. Normalization is omitted from most formulas for clarity.
  }
  \label{tab:residual_compare}
  \small
  \setlength{\tabcolsep}{8pt}
  \renewcommand{\arraystretch}{1.3}
  \resizebox{\textwidth}{!}{%
    \begin{tabular}{l l l c c}
      \toprule
      \multicolumn{2}{l}{\textbf{Method}}                                   & \textbf{Update rule}                                                                                                                                                                         & \textbf{Weight}                                                                                                           & \textbf{Source}                                                                  \\
      \midrule
      \multicolumn{5}{@{}l}{\emph{Single-state recurrence: layer $l$ receives only $\bm{h}_{l-1}$}}                                                                                                                                                                                                                                                                                                                                                                                       \\
      \addlinespace[2pt]
      \multicolumn{2}{l}{Residual \citep{he2015resnet}}                     & $\bm{h}_{l} = \bm{h}_{l-1} + f_{l-1}(\bm{h}_{l-1})$                                                                                                                                          & Fixed                                                                                                                     & $\bm{h}_{l-1}$                                                                   \\
      \multicolumn{2}{l}{ReZero \citep{bachlechner2020rezero}}              & $\bm{h}_{l} = \bm{h}_{l-1} + \alpha_{l} \cdot f_{l-1}(\bm{h}_{l-1})$                                                                                                                         & Static                                                                                                                    & $\bm{h}_{l-1}$                                                                   \\
      \multicolumn{2}{l}{LayerScale \citep{touvron2021going}}               & $\bm{h}_{l} = \bm{h}_{l-1} + \mathrm{diag}(\bm{\lambda}_{l}) \cdot f_{l-1}(\bm{h}_{l-1})$                                                                                                    & Static                                                                                                                    & $\bm{h}_{l-1}$                                                                   \\
      \multicolumn{2}{l}{Highway \citep{srivastava2015highway}}             & $\bm{h}_{l} = (1{-}\bm{g}_{l}) \odot \bm{h}_{l-1} + \bm{g}_{l} \odot f_{l-1}(\bm{h}_{l-1})$                                                                                                  & Dynamic                                                                                                                   & $\bm{h}_{l-1}$                                                                   \\
      \multicolumn{2}{l}{DeepNorm \citep{wang2022deepnorm}}                 & $\bm{h}_{l} = \mathrm{Norm}(\alpha \, \bm{h}_{l-1} + f_{l-1}(\bm{h}_{l-1}))$                                                                                                                 & Fixed                                                                                                                     & $\bm{h}_{l-1}$                                                                   \\
      \multicolumn{2}{l}{KEEL \citep{chen2026postlayernorm}}                & $\bm{h}_{l} = \mathrm{Norm}(\alpha \, \bm{h}_{l-1} + f_{l-1}(\mathrm{Norm}(\bm{h}_{l-1})))$                                                                                                  & Fixed                                                                                                                     & $\bm{h}_{l-1}$                                                                   \\
      \midrule
      \multicolumn{5}{@{}l}{\emph{Multi-state recurrence: layer $l$ receives $m$ streams}}                                                                                                                                                                                                                                                                                                                                                                                                \\
      \addlinespace[2pt]
      \multicolumn{2}{l}{SiameseNorm \citep{li2026siamesenorm}}             & $\bm{h}^{1}_{l} {=} \mathrm{Norm}(\bm{h}^{1}_{l-1} {+} \bm{y}_{l-1})$;\; $\bm{h}^{2}_{l} {=} \bm{h}^{2}_{l-1} {+} \bm{y}_{l-1}$                                                              & Fixed                                                                                                                     & 2 streams                                                                        \\
      \multicolumn{2}{l}{HC/mHC \citep{zhu2025hyperconnections,xie2026mhc}} & $\mathbf{H}_{l} = \mathbf{H}_{l-1}\mathbf{A}_{l} + f_{l-1}(\mathbf{H}_{l-1}\bm{\alpha}_{l-1})\,\bm{\beta}_{l-1}^\top$                                                                        & Dynamic                                                                                                                   & $m$ streams                                                                      \\
      \multicolumn{2}{l}{DDL \citep{zhang2026ddl}}                          & $\mathbf{H}_{l} = (\mathbf{I} {-} \beta_{l} \bm{k}_{l}\bm{k}_{l}^\top)\mathbf{H}_{l-1} + \beta_{l} \bm{k}_{l}\bm{v}_{l}^\top$                                                                & Dynamic                                                                                                                   & $d_v$ streams                                                                    \\
      \midrule
      \multicolumn{5}{@{}l}{\emph{Cross-layer access: layer $l$ can access individual earlier-layer outputs}}                                                                                                                                                                                                                                                                                                                                                                             \\
      \addlinespace[2pt]
      \multicolumn{2}{l}{DenseNet \citep{huang2018densenet}}                & $\bm{h}_{l} = \mathrm{ConvPool}([\bm{h}_1;\; f_1(\bm{h}_1);\; \ldots;\; f_{l-1}(\bm{h}_{l-1})])$                                                                                             & Static                                                                                                                    & $[\bm{h}_1,\ldots,\bm{h}_{l-1}]$                                                 \\
      \multicolumn{2}{l}{DenseFormer \citep{pagliardini2024denseformer}}    & $\bm{h}_{l} = \alpha_{0 \to l}\,\bm{h}_1 + \sum_{i=1}^{l-1} \alpha_{i \to l}\,f_i(\bm{h}_i)$                                                                                                 & Static                                                                                                                    & $[\bm{h}_1,\ldots,\bm{h}_{l-1}]$                                                 \\
      \multicolumn{2}{l}{MRLA \citep{fang2023cross}$^{1}$}                  & $\bm{h}_{l} = \sum_{i=1}^{l-1} \sigma\!\bigl(\mathrm{ConvPool}(f_{l-1}(\bm{h}_{l-1}))\bigr)^{\!\top} \sigma\!\bigl(\mathrm{ConvPool}(f_{i}(\bm{h}_i))\bigr) \, \mathrm{Conv}(f_i(\bm{h}_i))$ & Dynamic                                                                                                                   & $[\bm{h}_1,\ldots,\bm{h}_{l-1}]$                                                 \\
      \midrule
      \rowcolor{gray!10}
                                                                            & Full$^{2}$                                                                                                                                                                                   & $\bm{h}_{l} \propto \sum_{i=0}^{l-1} \phi(\bm{w}_{l}, \bm{k}_i) \, \bm{v}_i$                                              & Dynamic                          & $[\bm{h}_1,\ldots,\bm{h}_{l-1}]$              \\
      \rowcolor{gray!10}
      \multirow{-2}{*}{AttnRes (ours)}                                      & Block$^{3}$                                                                                                                                                                                  & $\bm{h}_{l} \propto \sum_{i=0}^{n-1} \phi(\bm{w}_{l}, \bm{k}_i) \, \bm{v}_i + \phi(\bm{w}_{l}, \bm{k}_n^j) \, \bm{v}_n^j$ & Dynamic                          & $[\bm{b}_0,\ldots,\bm{b}_{n{-}1},\bm{b}_n^j]$ \\
      \bottomrule
    \end{tabular}%
  }
  \par\vspace{4pt}
  \raggedright\footnotesize
  $^{1}$\,ConvPool: pooling operation followed by convolution (channel projection).\\
  $^{2}$\,$\phi(\bm{q},\bm{k}) = \exp\left(\bm{q}^\top \operatorname{RMSNorm}(\bm{k})\right)$;\; $\bm{k}_i = \bm{v}_i$;\; $\bm{v}_0 = \bm{h}_1$,\; $\bm{v}_{i\geq1} = f_i(\bm{h}_i)$. $\operatorname{softmax}$ jointly normalized over all sources.\\
  $^{3}$\,Same $\phi$ and normalization as Full;\; $\bm{v}_i = \bm{b}_i$,\; $\bm{v}_n^j = \bm{b}_n^j$.
\end{table*}

\subsection{Residual Connections as Structured Matrices}
\label{sec:structured-depth}

The residual variants discussed above can all be viewed as weighted aggregations over previous layer outputs.
We formalize this with a \emph{depth mixing matrix} $\mathbf{M} \in \mathbb{R}^{L \times L}$, where $\mathbf{M}_{i \to l}$ is the weight that layer $l$ assigns to the output of layer $i$.
The variants differ in how these weights arise (fixed, learned, or input-dependent) and whether $\mathbf{M}$ is constrained to low rank or allowed to be dense.
The semiseparable rank of $\mathbf{M}$~\cite{mamba2} offers a unified lens for comparing them.

\begin{figure*}[t]
    \centering
    \scriptsize

    \definecolor{CellBlue}{HTML}{E1EBF5}   
    \definecolor{CellSand}{HTML}{F5E8DF}   
    \definecolor{CellPurple}{HTML}{EBE4F0} 
    \definecolor{CellMint}{HTML}{E1F0E5}   
    \definecolor{CellBlush}{HTML}{F5E1E6}  

    \renewcommand{\arraystretch}{1.4}
    \newlength{\gridw}\setlength{\gridw}{0.49\textwidth}
    \newlength{\cellw}\setlength{\cellw}{6.5em}
    \newcommand{\cell}[1]{\makebox[\cellw][c]{$#1$}}

    \newcommand{\roundcell}[3]{%
        \fill[#3, rounded corners=2pt] (#1-|#2) rectangle (\the\numexpr#1+1\relax-|\the\numexpr#2+1\relax);
    }

    \makebox[\gridw][c]{%
        \begin{tabular}{@{}c@{}}
            Highway \\[\medskipamount]
            $\begin{bNiceArray}{llll}[margin=2pt]
                     \cell{1}                         & \cell{}                               & \cell{}                               & \cell{}    \\
                     \cell{\gamma_{1 \to 2}^{\times}} & \cell{g_2}                            & \cell{}                               & \cell{}    \\
                     \cell{\gamma_{1 \to 3}^{\times}} & \cell{g_2\,\gamma_{2 \to 3}^{\times}} & \cell{g_3}                            & \cell{}    \\
                     \cell{\gamma_{1 \to 4}^{\times}} & \cell{g_2\,\gamma_{2 \to 4}^{\times}} & \cell{g_3\,\gamma_{3 \to 4}^{\times}} & \cell{g_4}
                 \end{bNiceArray}$
        \end{tabular}%
    }%
    \hfill
    \makebox[\gridw][c]{%
        \begin{tabular}{@{}c@{}}
            (m)HC \\[\medskipamount]
            $\begin{bNiceArray}{llll}[margin=2pt]
                     \cell{\bm{\beta}_0^\top \bm{\alpha}_1}                                 & \cell{}                                                                & \cell{}                                                              & \cell{}                                \\
                     \cell{\bm{\beta}_0^\top \mathbf{A}_{1 \to 2}^{\times} \bm{\alpha}_2}   & \cell{\bm{\beta}_1^\top \bm{\alpha}_2}                                 & \cell{}                                                              & \cell{}                                \\
                     \cell{\bm{\beta}_0^\top \mathbf{A}_{1 \to 3}^{\times}\, \bm{\alpha}_3} & \cell{\bm{\beta}_1^\top \mathbf{A}_{2 \to 3}^{\times} \bm{\alpha}_3}   & \cell{\bm{\beta}_2^\top \bm{\alpha}_3}                               & \cell{}                                \\
                     \cell{\bm{\beta}_0^\top \mathbf{A}_{1 \to 4}^{\times}\, \bm{\alpha}_4} & \cell{\bm{\beta}_1^\top \mathbf{A}_{2 \to 4}^{\times}\, \bm{\alpha}_4} & \cell{\bm{\beta}_2^\top \mathbf{A}_{3 \to 4}^{\times} \bm{\alpha}_4} & \cell{\bm{\beta}_3^\top \bm{\alpha}_4}
                 \end{bNiceArray}$
        \end{tabular}%
    }

    \vspace{14pt}

    \makebox[\gridw][c]{%
        \begin{tabular}{@{}c@{}}
            Full AttnRes \\[\medskipamount]
            $\begin{bNiceArray}{llll}[margin=2pt]
                     \CodeBefore
                     \begin{tikzpicture}
                        \roundcell{1}{1}{CellBlue}
                        \roundcell{2}{1}{CellBlue}
                        \roundcell{3}{1}{CellBlue}
                        \roundcell{4}{1}{CellBlue}
                        \roundcell{2}{2}{CellSand}
                        \roundcell{3}{2}{CellSand}
                        \roundcell{4}{2}{CellSand}
                        \roundcell{3}{3}{CellPurple}
                        \roundcell{4}{3}{CellPurple}
                        \roundcell{4}{4}{CellMint}
                    \end{tikzpicture}
                     \Body
                     \cell{\phi\left(\bm{w}_1, \bm{k}_0\right)} & \cell{}                              & \cell{}                            & \cell{}                            \\
                     \cell{\phi\left(\bm{w}_2, \bm{k}_0\right)} & \cell{\phi\left(\bm{w}_2, \bm{k}_1\right)} & \cell{}                            & \cell{}                            \\
                     \cell{\phi\left(\bm{w}_3, \bm{k}_0\right)} & \cell{\phi\left(\bm{w}_3, \bm{k}_1\right)} & \cell{\phi\left(\bm{w}_3, \bm{k}_2\right)} & \cell{}                            \\
                     \cell{\phi\left(\bm{w}_4, \bm{k}_0\right)} & \cell{\phi\left(\bm{w}_4, \bm{k}_1\right)} & \cell{\phi\left(\bm{w}_4, \bm{k}_2\right)} & \cell{\phi\left(\bm{w}_4, \bm{k}_3\right)}
                 \end{bNiceArray}$
        \end{tabular}%
    }%
    \hfill
    \makebox[\gridw][c]{%
        \begin{tabular}{@{}c@{}}
            Block AttnRes \\[\medskipamount]
            $\begin{bNiceArray}{llll}[margin=2pt]
                     \CodeBefore
                     \begin{tikzpicture}
                        \roundcell{1}{1}{CellBlue}
                        \roundcell{2}{1}{CellBlue}
                        \roundcell{3}{1}{CellBlue}
                        \roundcell{4}{1}{CellBlue}
                        \roundcell{2}{2}{CellSand}
                        \draw[dashed, rounded corners=2pt, line width=0.4pt] (2-|2) rectangle (3-|3);
                        \fill[CellBlush, rounded corners=2pt] (3-|2) rectangle (4-|4);
                        \fill[CellBlush, rounded corners=2pt] (4-|2) rectangle (5-|4);
                        \roundcell{4}{4}{CellMint}
                        \draw[dashed, rounded corners=2pt, line width=0.4pt] (4-|4) rectangle (5-|5);
                    \end{tikzpicture}
                     \Body
                     \cell{\phi\left(\bm{w}_1, \bm{k}_0\right)} & \cell{}                                  & \cell{}                            & \cell{}                                \\
                     \cell{\phi\left(\bm{w}_2, \bm{k}_0\right)} & \cell{\phi\left(\bm{w}_2, \bm{k}_1\right)} & \cell{}                            & \cell{}                                \\
                     \cell{\phi\left(\bm{w}_3, \bm{k}_0\right)} & \Block{1-2}{\cell{\phi\left(\bm{w}_3, \bm{k}_1+\bm{k}_2\right)}} &                            & \cell{}                                \\
                     \cell{\phi\left(\bm{w}_4, \bm{k}_0\right)} & \Block{1-2}{\cell{\phi\left(\bm{w}_4, \bm{k}_1 +\bm{k}_2\right)}} &                            & \cell{\phi\left(\bm{w}_4, \bm{k}_3\right)}
                 \end{bNiceArray}$
        \end{tabular}%
    }

    \caption{Depth mixing matrices $\mathbf{M}$ for four residual variants ($L{=}4$; Block AttnRes uses block size $S{=}2$). Highway is shown with scalar gates for clarity. AttnRes panels show unnormalized $\phi$ scores; background colors group entries that share the same source (Full AttnRes) or the same source block (Block AttnRes).}
    \label{fig:depth-matrices}
\end{figure*}
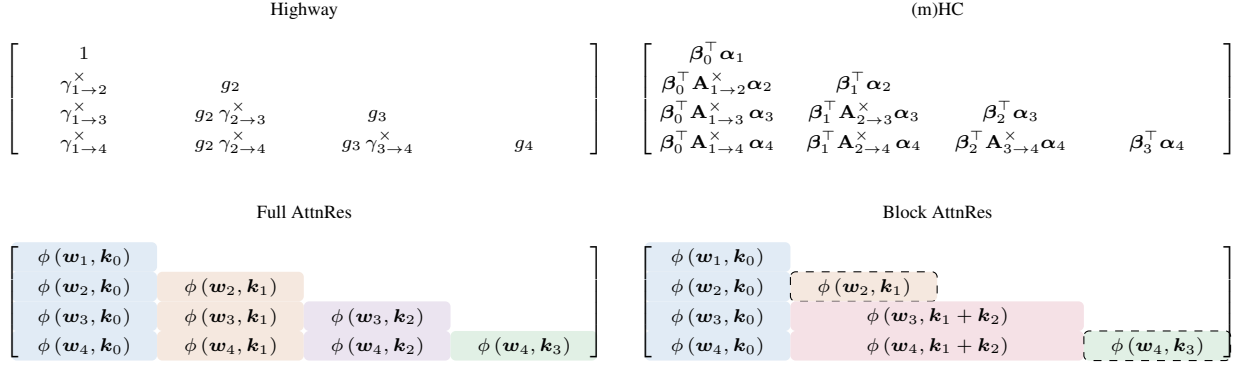

Concretely, the input to layer $l$ is $\bm{h}_l = \sum_{i=0}^{l-1} \mathbf{M}_{i \to l}\, \bm{v}_i$, where $\bm{v}_0 = \bm{h}_1$ (embedding) and $\bm{v}_i = f_i(\bm{h}_i)$ for $i \geq 1$.
Fig.~\ref{fig:depth-matrices} visualizes $\mathbf{M}$ for representative methods; we derive each below.
\begin{itemize}[leftmargin=*]
    \item Standard residual~\cite{he2015resnet}, $\bm{h}_{l} = \bm{h}_{l-1} + f_{l-1}(\bm{h}_{l-1})$. Expanding gives $\bm{h}_l = \sum_{i=0}^{l-1} \bm{v}_i$, so $\mathbf{M}_{i \to l} = 1$ for all $i < l$ and $\mathbf{M}$ is an all-ones lower-triangular matrix:
          \[
              \begin{bmatrix}
                  \bm{h}_1 \\ \bm{h}_2 \\ \vdots \\ \bm{h}_L
              \end{bmatrix}
              =
              \begin{bmatrix}
                  1      &        &        &   \\
                  1      & 1      &        &   \\
                  \vdots & \vdots & \ddots &   \\
                  1      & 1      & \cdots & 1
              \end{bmatrix}
              \begin{bmatrix}
                  \bm{v}_0 \\ \bm{v}_1 \\ \vdots \\ \bm{v}_{L-1}
              \end{bmatrix}
          \]
    \item Highway~\cite{srivastava2015highway}, $\bm{h}_{l} = (1{-}g_{l})\,\bm{h}_{l-1} + g_{l}\,f_{l-1}(\bm{h}_{l-1})$ (written here with scalar gates for clarity; the element-wise extension is straightforward). Defining the carry product $\gamma_{i \to l}^{\times} \coloneqq \prod_{j=i+1}^{l}(1-g_j)$, the weights are $\mathbf{M}_{0 \to l} = \gamma_{1 \to l}^{\times}$ for the embedding and $\mathbf{M}_{i \to l} = g_{i+1}\,\gamma_{i+1 \to l}^{\times}$ for $i \geq 1$. Since the cumulative products factor through scalar gates, $\mathbf{M}$ is 1-semiseparable~\cite{mamba2}, the same rank as the standard residual but with input-dependent weights. The weights sum to one by construction, making Highway a softmax-free depth-wise instance of stick-breaking attention~\cite{tan2025stickbreaking}.
    \item (m)HC~\cite{zhu2025hyperconnections,xie2026mhc} maintain $m$ parallel streams $\mathbf{H}_{l} \in \mathbb{R}^{d \times m}$, updated via
          \[
              \mathbf{H}_{l} = \mathbf{H}_{l-1} \mathbf{A}_{l} + f_{l-1}(\mathbf{H}_{l-1} \bm{\alpha}_{l-1})\, \bm{\beta}_{l-1}^\top,
          \]
          where $\mathbf{A}_{l} \in \mathbb{R}^{m \times m}$ is a learned transition matrix, $\bm{\alpha}_{l-1} \in \mathbb{R}^{m}$ mixes streams into a single input for $f_{l-1}$, and $\bm{\beta}_{l-1} \in \mathbb{R}^{m}$ distributes the output back across streams.
          Unrolling the recurrence gives the effective weight
          \begin{equation}
              \label{eq:hc-unroll}
              \mathbf{M}_{i \to l} = \bm{\beta}_{i}^\top \, \mathbf{A}_{i+1 \to l}^{\times} \, \bm{\alpha}_{l},
          \end{equation}
          where $\mathbf{A}_{i \to j}^{\times} \coloneqq \prod_{k=i+1}^{j} \mathbf{A}_k$.
          The $m \times m$ transitions render $\mathbf{M}$ $m$-semiseparable~\cite{mamba2}. mHC~\cite{xie2026mhc,yang2026mhclite} further constrains each $\mathbf{A}_l$ to be doubly stochastic, stabilizing the cumulative products across depth.
    \item Full AttnRes computes $\mathbf{M}_{i \to l} = \brickred{\alpha_{i \to l}}$ via $\phi(\bm{w}_l, \bm{k}_i) = \exp\left(\bm{w}_l^\top \operatorname{RMSNorm}(\bm{k}_i)\right)$ with normalization, where $\bm{k}_i = \bm{v}_i$ are input-dependent layer outputs, yielding a dense, rank-$L$ $\mathbf{M}$.
    \item Block AttnRes partitions layers into $N$ blocks $\mathcal{B}_1, \ldots, \mathcal{B}_N$. For sources $i$ in a completed earlier block $\mathcal{B}_n$, all share the block-level key/value $\bm{b}_n$, so $\mathbf{M}_{i \to l} = \brickred{\alpha_{n \to l}}$ for every $i \in \mathcal{B}_n$. Within the current block, each layer additionally attends over the evolving partial sum $\bm{b}_n^{i-1}$, introducing one extra distinct source per intra-block position. The effective rank of $\mathbf{M}$ therefore lies between $N$ and $N + S$ (where $S$ is the block size), interpolating between standard residual ($N{=}1$) and Full AttnRes ($N{=}L$).
\end{itemize}

\paragraph{Practicality.}
The structured-matrix perspective serves two purposes.
First, it enables analytical insights that are not apparent from the recurrence form alone.
The input-dependent $\mathbf{M}$ of AttnRes, for instance, reveals depth-wise attention sinks (\S\ref{sec:analysis}), where certain layers consistently attract high weight regardless of input, mirroring the same phenomenon in sequence-wise attention~\cite{xiao2023efficient}.
Second, it informs new designs by exposing which properties of the kernel $\phi$ matter.
For example, when $\phi$ decomposes as $\phi(\bm{q}, \bm{k}) = \varphi(\bm{q})^\top \varphi(\bm{k})$ for some feature map $\varphi$~\cite{katharopoulos-2020-transformers}, depth-wise attention collapses into a recurrence---precisely the structure underlying the MRLA--GLA and DDL--DeltaNet correspondences noted above.

\paragraph{Prior Residuals as Depth-Wise Linear Attention}

The structured-matrix perspective further relates to the sequence-depth duality by showing that existing residual variants are, in effect, instances of \emph{linear} attention over the depth axis.
For example, the unrolled (m)HC weight $\mathbf{M}_{i \to l} = \boldsymbol{\beta}_i^\top \mathbf{A}_{i+1 \to l}^\times \boldsymbol{\alpha}_l$ (Eq.~\ref{eq:hc-unroll}) admits a natural attention interpretation in which $\boldsymbol{\alpha}_l$ plays the role of a query issued by layer $l$, $\boldsymbol{\beta}_i$ serves as a key summarizing the contribution of layer $i$, and the cumulative transition $\mathbf{A}_{i+1 \to l}^\times$ acts as a depth-relative positional operator~\cite{zhang2025kimi} governing the query--key interaction across intervening layers.
Notably, the $m$ parallel streams correspond to state expansion~\cite{qin-2024-hgrn2,mak2025residualmatrix} along the depth axis, expanding the recurrent state from $d$ to $d {\times}\, m$ and thereby increasing the semiseparable rank of $\mathbf{M}$.
\cite{xie2026mhcidentity} show that replacing $\mathbf{A}_{i+1 \to l}^{\times}$ with the identity matrix still yields competitive performance, highlighting the role of state expansion.
Through this lens, methods like (m)HC thus act as depth-wise \emph{linear} attention with matrix-valued states, while AttnRes acts as depth-wise $\operatorname{softmax}$ attention.

\section{Related Work}

\paragraph{Normalization, Scaling, and Depth Stability.}
The standard residual update $\bm{h}_{l+1} = \bm{h}_l + f_l(\bm{h}_l)$~\citep{he2015resnet} presents a fundamental tension between \emph{normalization placement} and \emph{gradient propagation}.
PostNorm~\citep{vaswani-2017-attention} maintains bounded magnitudes but distorts gradients, as repeated normalization on the residual path compounds into gradient vanishing at depth~\citep{xiong2020layer}.
PreNorm~\citep{nguyen-salazar-2019-transformers,xiong2020layer} restores a clean identity path yet introduces unbounded magnitude growth: since $\|\bm{h}_l\|$ grows as $O(L)$, each layer's relative contribution shrinks, compelling deeper layers to produce ever-larger outputs and limiting effective depth~\citep{li2026siamesenorm}.
Subsequent work reconciles both desiderata via scaled residual paths~\citep{wang2022deepnorm}, hybrid normalization~\citep{zhuo2025hybridnorm}, amplified skip connections~\citep{chen2026postlayernorm}, or learned element-wise gates~\citep{srivastava2015highway} (see Table~\ref{tab:residual_compare}).
AttnRes sidesteps this tension by replacing the additive recurrence with selective aggregation over individual earlier-layer outputs, avoiding both the cumulative magnitude growth of PreNorm and the repeated scale contraction of PostNorm.

\paragraph{Multi-State Recurrence.}
All single-state methods above condition layer $l$ only on $\bm{h}_{l-1}$, from which individual earlier-layer contributions cannot be selectively retrieved.
Several methods address this by widening the recurrence to multiple parallel streams:
Hyper-Connections~\citep{zhu2025hyperconnections} and its stabilized variant mHC~\citep{xie2026mhc} maintain $m$ streams with learned mixing matrices;
DDL~\citep{zhang2026ddl} maintains a matrix state updated via a delta-rule erase-and-write mechanism;
SiameseNorm~\citep{li2026siamesenorm} maintains two parameter-shared streams---one PreNorm and one PostNorm---to preserve identity gradients and bounded representations.
While these methods alleviate information compression, they still condition on the immediate predecessor's state; AttnRes is orthogonal, providing selective access to individual earlier-layer outputs while remaining compatible with any normalization or gating scheme.
We discuss the formal connection to Hyper-Connections in \S~\ref{sec:structured-depth}.

\paragraph{Cross-Layer Connectivity.}
A separate line of work bypasses the single-state bottleneck by giving each layer direct access to individual earlier-layer outputs.
The simplest approach uses static weights: DenseNet~\citep{huang2018densenet} concatenates all preceding feature maps; ELMo~\citep{peters-etal-2018-deep} computes a $\operatorname{softmax}$-weighted sum of layer representations with learned scalar weights; DenseFormer~\citep{pagliardini2024denseformer} and ANCRe~\citep{zhang2026ancre} assign learned per-layer scalar coefficients fixed after training.
For input-dependent aggregation, MUDDFormer~\citep{xiao2025muddformer} generates position-dependent weights via a small MLP across four decoupled streams;
MRLA~\citep{fang2023cross} applies element-wise sigmoid gating over all previous layers, though its separable query--key product is closer to linear attention than $\operatorname{softmax}$-based retrieval.
Other methods trade full cross-layer access for more targeted designs:
Value Residual Learning~\citep{zhou-etal-2025-value} accesses only a single earlier layer;
LAuReL~\citep{menghani2025laurel} augments the residual with low-rank projections over the previous $k$ activations;
Dreamer~\citep{knupp2026depthrecurrent} combines sequence attention with depth attention and sparse experts.
AttnRes combines $\operatorname{softmax}$-normalized, input-dependent weights with selective access to all preceding layers through a single $d$-dimensional pseudo-query per layer, and introduces a block structure reducing cost from $O(L^2)$ to $O(LN)$.
Cache-based pipeline communication and a two-phase computation strategy (\S~\ref{sec:infra}) make Block AttnRes practical at scale with negligible overhead.
\section*{Conclusion}

Inspired by the duality between sequence and depth, we introduce AttnRes, which replaces fixed, uniform residual accumulation with learned, input-dependent depth-wise attention. We validate the method through ablation studies and scaling law experiments, showing that its gains persist across scales. Because Full AttnRes must access all preceding layer outputs at every layer, the memory footprint of cross-layer aggregation grows as $O(Ld)$, which is prohibitive for large-scale models on current hardware. We therefore introduce Block AttnRes, which partitions layers into $N$ blocks and attends over block-level representations. Empirically, using about 8 blocks recovers most of the gains of Full AttnRes, while finer-grained blocking remains a promising direction as future hardware constraints relax. Together with cross-stage caching and a two-phase computation strategy, Block AttnRes is practical at scale, incurring only marginal training overhead and minimal inference overhead.

\newpage
\printbibliography[title={References}]

\newpage
\appendix
\section{Contributions}

The authors are listed in order of the significance of their contributions, with those in project leadership roles appearing last.

\begin{multicols}{1}

Guangyu Chen$^*$\\
Yu Zhang$^*$\\
Jianlin Su$^*$\\
Weixin Xu\\
Siyuan Pan\\
Yaoyu Wang\\
Yucheng Wang\\
Guanduo Chen\\
Bohong Yin\\
Yutian Chen\\
Junjie Yan\\
Ming Wei\\
Y. Zhang\\
Fanqing Meng\\
Chao Hong\\
Xiaotong Xie\\
Shaowei Liu\\
Enzhe Lu\\
Yunpeng Tai\\
Yanru Chen\\
Xin Men\\
Haiqing Guo\\
Y. Charles\\
Haoyu Lu\\
Lin Sui\\
Jinguo Zhu\\
Zaida Zhou\\
Weiran He\\
Weixiao Huang\\
Xinran Xu\\
Yuzhi Wang\\
Guokun Lai\\
Yulun Du\\
Yuxin Wu\\
Zhilin Yang\\
Xinyu Zhou\\

\end{multicols}

$^*$ Equal contribution

\newpage
\section{Optimized Inference I/O for Full Attention Residuals}
\label{app:inference}

A na\"ive implementation of Full AttnRes scans all preceding layer outputs at every layer, so memory traffic scales linearly with depth. As noted in \S\ref{sec:infra-inference}, however, the pseudo-query $\bm{w}_l$ is a learned parameter independent of both the input and the hidden state. We can therefore batch inter-block accesses across layers in a two-phase schedule, bringing total I/O well below the na\"ive bound.

Note that the block partition introduced below is purely an inference scheduling device. Unlike Block AttnRes, it leaves the model architecture unchanged and does not replace per-layer sources with block summaries; it simply makes the amortization argument concrete.

\paragraph{Setup}

Let the model have $L$ layers and hidden dimension $d$, partitioned into $N$ contiguous blocks of size $S = L/N$. Inference proceeds one block at a time: Phase~1 jointly computes inter-block attention for all $S$ layers in the block against all preceding blocks, and Phase~2 walks through intra-block dependencies sequentially.

\subsection*{Phase 1: Batched Inter-block Attention}

Consider block $n$ with its $S$ layers. The queries $\{\bm{w}_l\}_{l \in \mathcal{B}_n}$ are all known before execution begins, so the $(n{-}1)S$ preceding key--value pairs need only be read once from HBM and reused across all $S$ queries. The read cost for block $n$ is therefore
\begin{equation}
    \mathrm{Read}_{\text{inter}}^{(n)} = 2(n-1)Sd,
\end{equation}
where the factor of $2$ accounts for both keys and values. Summing over all $N$ blocks and using $SN=L$:
\begin{equation}
    \mathrm{Read}_{\text{inter}} = \sum_{n=1}^{N} 2(n-1)Sd
    = 2Sd \cdot \frac{N(N-1)}{2}
    = dL(N-1).
\end{equation}

Phase~1 also writes one $d$-dimensional output per layer, giving $\mathrm{Write}_{\text{inter}}^{(n)} = Sd$ per block and
\begin{equation}
    \mathrm{Write}_{\text{inter}} = Ld
\end{equation}
in total.

\subsection*{Phase 2: Sequential Intra-block Attention}

Phase~1 covers all sources before the current block. Within the block, however, each layer depends on those before it, so these must be handled in order. Layer $t$ ($1 \le t \le S$) reads $t{-}1$ intra-block key--value pairs at a cost of $2(t{-}1)d$. Summing over one block:
\begin{equation}
    \mathrm{Read}_{\text{intra}}^{(n)} = \sum_{t=1}^{S} 2(t-1)d = S(S-1)d.
\end{equation}
Phase~2 also writes one output per layer, so $\mathrm{Write}_{\text{intra}}^{(n)} = Sd$.

\subsection*{Total Amortized I/O per Layer}

Summing both phases over all $N$ blocks:
\begin{equation}
    \mathrm{Read}_{\text{total}} = dL(N-1) + N \cdot S(S-1)d, \qquad
    \mathrm{Write}_{\text{total}} = 2Ld.
\end{equation}
Dividing by $L$ and using $SN=L$:
\begin{equation}
    \text{Read per layer} = (N-1)d + (S-1)d = (S + N - 2)d, \qquad
    \text{Write per layer} = 2d,
\end{equation}
\begin{equation}
    \boxed{\;\text{Total I/O per layer} = (S + N)\,d.\;}
\end{equation}

Batching inter-block reads thus brings per-layer I/O from $\mathcal{O}(L)$ down to $\mathcal{O}(S{+}N)$. The schedule follows the same two-phase split as Block AttnRes: inter-block attention accounts for the bulk of the traffic, while sequential computation stays local within each block.

\end{document}